\documentclass[10pt,twocolumn,letterpaper]{article}

\usepackage[margin=0.7in,columnsep=0.25in]{geometry}
\usepackage[T1]{fontenc}
\usepackage[utf8]{inputenc}
\usepackage{newtxtext,newtxmath}
\usepackage{microtype}
\usepackage{graphicx}
\usepackage{booktabs}
\usepackage{array}
\usepackage{needspace}
\usepackage{amsmath}
\usepackage{xcolor}
\usepackage[ruled,vlined,linesnumbered]{algorithm2e}
\usepackage[font=small,labelfont=bf,skip=4pt]{caption}
\usepackage{subcaption}
\usepackage{placeins}
\usepackage[numbers,sort&compress]{natbib}
\usepackage{url}
\usepackage[colorlinks=true,allcolors=black,breaklinks=true]{hyperref}
\hypersetup{pdftitle={Human Demonstrations as Post-Training Data for Any Robot Body: Retargeting the Object, Not the Hand},pdfauthor={Nagasrikanth Kallakuri}}

\graphicspath{{figures/}}

\makeatletter
\renewcommand\paragraph{\@startsection{paragraph}{4}{\z@}{1.2ex \@plus 0.2ex \@minus 0.1ex}{-0.6em}{\normalfont\normalsize\bfseries}}%
\makeatother

\providecommand{\TierLine}{}\renewcommand{\TierLine}{kinematic feasibility $\cdot$ $\pm$50\,mm / $\pm$15$^\circ$ $\cdot$ in simulation}
\providecommand{\ResultsGeneratedUTC}{}\renewcommand{\ResultsGeneratedUTC}{2026-10-14T00:31:58Z}
\providecommand{\CorpusCaptures}{}\renewcommand{\CorpusCaptures}{42}
\providecommand{\CorpusSolveRuns}{}\renewcommand{\CorpusSolveRuns}{45}
\providecommand{\CorpusBodies}{}\renewcommand{\CorpusBodies}{8}
\providecommand{\CorpusBodyClasses}{}\renewcommand{\CorpusBodyClasses}{4}
\providecommand{\CorpusPairs}{}\renewcommand{\CorpusPairs}{157}
\providecommand{\CorpusCaptureMinutes}{}\renewcommand{\CorpusCaptureMinutes}{19.2}
\providecommand{\ConstFrameRate}{}\renewcommand{\ConstFrameRate}{10\,Hz}
\providecommand{\CorpusVerifiedRobotMinutes}{}\renewcommand{\CorpusVerifiedRobotMinutes}{6.1}
\providecommand{\OutcomePass}{}\renewcommand{\OutcomePass}{49}
\providecommand{\OutcomePassShare}{}\renewcommand{\OutcomePassShare}{31.2\%}
\providecommand{\OutcomeFail}{}\renewcommand{\OutcomeFail}{82}
\providecommand{\OutcomeFailShare}{}\renewcommand{\OutcomeFailShare}{52.2\%}
\providecommand{\OutcomeRefused}{}\renewcommand{\OutcomeRefused}{22}
\providecommand{\OutcomeRefusedShare}{}\renewcommand{\OutcomeRefusedShare}{14.0\%}
\providecommand{\OutcomeUnauditable}{}\renewcommand{\OutcomeUnauditable}{4}
\providecommand{\OutcomeUnauditableShare}{}\renewcommand{\OutcomeUnauditableShare}{2.5\%}
\providecommand{\OutcomeAttempted}{}\renewcommand{\OutcomeAttempted}{131}
\providecommand{\OutcomePassShareOfAttempted}{}\renewcommand{\OutcomePassShareOfAttempted}{37.4\%}
\providecommand{\BodyLowCostFiveDofPairs}{}\renewcommand{\BodyLowCostFiveDofPairs}{1}
\providecommand{\FleetComposition}{}\renewcommand{\FleetComposition}{four bodies with seven joints, three with six and one with five; one body has no closing joint}
\providecommand{\RefusalVocabularySize}{}\renewcommand{\RefusalVocabularySize}{7}
\providecommand{\ContactWitnessCount}{}\renewcommand{\ContactWitnessCount}{3}
\providecommand{\FailureCollisionShare}{}\renewcommand{\FailureCollisionShare}{52.4\%}
\providecommand{\RefusalReasonsDistinct}{}\renewcommand{\RefusalReasonsDistinct}{5}
\providecommand{\RefusalTopReasonShare}{}\renewcommand{\RefusalTopReasonShare}{50.0\%}
\providecommand{\RefusalNoActuatedGripper}{}\renewcommand{\RefusalNoActuatedGripper}{11}
\providecommand{\FailureTopMode}{}\renewcommand{\FailureTopMode}{collision self}
\providecommand{\FailureTopModeShare}{}\renewcommand{\FailureTopModeShare}{23.2\%}
\providecommand{\ScoredFracSpanMedian}{}\renewcommand{\ScoredFracSpanMedian}{79.1\%}
\providecommand{\ScoredFracTakeMedian}{}\renewcommand{\ScoredFracTakeMedian}{15.9\%}
\providecommand{\ContactScoredFrames}{}\renewcommand{\ContactScoredFrames}{3\,517}
\providecommand{\ContactCoMotionShare}{}\renewcommand{\ContactCoMotionShare}{6.7\%}
\providecommand{\ContactPairsWithEvidence}{}\renewcommand{\ContactPairsWithEvidence}{79}
\providecommand{\RetractedPairs}{}\renewcommand{\RetractedPairs}{35}
\providecommand{\RetractedCaptures}{}\renewcommand{\RetractedCaptures}{16}
\providecommand{\RetractedPairShare}{}\renewcommand{\RetractedPairShare}{22.3\%}
\providecommand{\RetractionCausePreconditionMeasured}{}\renewcommand{\RetractionCausePreconditionMeasured}{10}
\providecommand{\RetractionCauseArtifactReminted}{}\renewcommand{\RetractionCauseArtifactReminted}{25}
\providecommand{\RetractionCauseNewVerifierCheck}{}\renewcommand{\RetractionCauseNewVerifierCheck}{0}
\providecommand{\ContractRefusals}{}\renewcommand{\ContractRefusals}{34}
\providecommand{\ContractBodyOnlyRefusalShare}{}\renewcommand{\ContractBodyOnlyRefusalShare}{76.5\%}
\providecommand{\ScopePairs}{}\renewcommand{\ScopePairs}{157}
\providecommand{\ScopeContractRows}{}\renewcommand{\ScopeContractRows}{146}
\providecommand{\ScopeScoredWindow}{}\renewcommand{\ScopeScoredWindow}{144}
\providecommand{\CandidateExtraRefusedOne}{}\renewcommand{\CandidateExtraRefusedOne}{43}
\providecommand{\CandidateExtraRefusedTwo}{}\renewcommand{\CandidateExtraRefusedTwo}{53}
\providecommand{\CandidateExtraRefusedThree}{}\renewcommand{\CandidateExtraRefusedThree}{54}
\providecommand{\ConstTubePos}{}\renewcommand{\ConstTubePos}{50\,mm}
\providecommand{\ConstTubeOri}{}\renewcommand{\ConstTubeOri}{15$^\circ$}
\providecommand{\ConstPassCoverage}{}\renewcommand{\ConstPassCoverage}{85\%}
\providecommand{\ConstFrictionMu}{}\renewcommand{\ConstFrictionMu}{0.40}
\providecommand{\ConstFrictionConeHalfAngle}{}\renewcommand{\ConstFrictionConeHalfAngle}{21.8$^\circ$}
\providecommand{\ConstClearanceGate}{}\renewcommand{\ConstClearanceGate}{5\,mm}
\providecommand{\ConstContactEnter}{}\renewcommand{\ConstContactEnter}{8\,mm}
\providecommand{\ConstContactRelease}{}\renewcommand{\ConstContactRelease}{20\,mm}
\providecommand{\ConstContactDwell}{}\renewcommand{\ConstContactDwell}{3}
\providecommand{\ConstCoMotionTol}{}\renewcommand{\ConstCoMotionTol}{30\,mm}
\providecommand{\ConstCoMotionWindow}{}\renewcommand{\ConstCoMotionWindow}{5}
\providecommand{\ConstCoMotionDisplaced}{}\renewcommand{\ConstCoMotionDisplaced}{50\,mm}
\providecommand{\PilotTakes}{}\renewcommand{\PilotTakes}{4}
\providecommand{\PilotProcessed}{}\renewcommand{\PilotProcessed}{4}
\providecommand{\PilotTierK}{}\renewcommand{\PilotTierK}{0}
\providecommand{\PilotGate}{}\renewcommand{\PilotGate}{0}
\providecommand{\PilotTask}{}\renewcommand{\PilotTask}{a marker picked off the bench and dropped into a container}
\providecommand{\PilotFailureModes}{}\renewcommand{\PilotFailureModes}{left the tube on three takes and self-collision on one take}
\providecommand{\PilotVisualCoverageContactMax}{}\renewcommand{\PilotVisualCoverageContactMax}{31\%}
\providecommand{\AblationEpisodes}{}\renewcommand{\AblationEpisodes}{42}
\providecommand{\AblationBodies}{}\renewcommand{\AblationBodies}{2}
\providecommand{\AblationTargetSourceCovControl}{}\renewcommand{\AblationTargetSourceCovControl}{100.0\%}
\providecommand{\AblationTargetSourceCovArm}{}\renewcommand{\AblationTargetSourceCovArm}{18.7\%}
\providecommand{\AblationGraspRankingPassControl}{}\renewcommand{\AblationGraspRankingPassControl}{33\%}
\providecommand{\AblationGraspRankingPassArm}{}\renewcommand{\AblationGraspRankingPassArm}{13\%}
\providecommand{\AblationGraspRankingUnchanged}{}\renewcommand{\AblationGraspRankingUnchanged}{77\%}
\providecommand{\AblationContactIntervalsPairs}{}\renewcommand{\AblationContactIntervalsPairs}{55}
\providecommand{\AblationContactIntervalsEpisodes}{}\renewcommand{\AblationContactIntervalsEpisodes}{40}
\providecommand{\AblationContactIntervalsPassArmNum}{}\renewcommand{\AblationContactIntervalsPassArmNum}{25}
\providecommand{\AblationPlacementPassControl}{}\renewcommand{\AblationPlacementPassControl}{33\%}
\providecommand{\AblationPlacementPassArm}{}\renewcommand{\AblationPlacementPassArm}{31\%}
\providecommand{\AblationOnsetHeldNum}{}\renewcommand{\AblationOnsetHeldNum}{40}
\providecommand{\AblationOnsetHeldDen}{}\renewcommand{\AblationOnsetHeldDen}{40}
\providecommand{\RefusalScopeRows}{}\renewcommand{\RefusalScopeRows}{146}
\providecommand{\RefusalEnforcedRestPose}{}\renewcommand{\RefusalEnforcedRestPose}{21}
\providecommand{\RefusalPrecisionKNum}{}\renewcommand{\RefusalPrecisionKNum}{13}
\providecommand{\RefusalPrecisionKDen}{}\renewcommand{\RefusalPrecisionKDen}{13}
\providecommand{\RefusalPrecisionKCI}{}\renewcommand{\RefusalPrecisionKCI}{77.2--100.0\%}
\providecommand{\RefusalPrecisionNoClosingJointDen}{}\renewcommand{\RefusalPrecisionNoClosingJointDen}{8}
\providecommand{\RefusalPrecisionWithClosureNum}{}\renewcommand{\RefusalPrecisionWithClosureNum}{5}
\providecommand{\RefusalPrecisionWithClosureDen}{}\renewcommand{\RefusalPrecisionWithClosureDen}{5}
\providecommand{\RefusalPrecisionWithClosureCI}{}\renewcommand{\RefusalPrecisionWithClosureCI}{56.6--100.0\%}
\providecommand{\RefusalPrecisionNoClosingJointWeldedNum}{}\renewcommand{\RefusalPrecisionNoClosingJointWeldedNum}{5}
\providecommand{\RefusalPrecisionNoClosingJointOtherNum}{}\renewcommand{\RefusalPrecisionNoClosingJointOtherNum}{3}
\providecommand{\RefusalBestCandidateRecallNum}{}\renewcommand{\RefusalBestCandidateRecallNum}{32}
\providecommand{\RefusalBestCandidateRecallDen}{}\renewcommand{\RefusalBestCandidateRecallDen}{63}
\providecommand{\RefusalBestCandidateRecall}{}\renewcommand{\RefusalBestCandidateRecall}{50.8\%}
\providecommand{\RefusalBestCandidateCost}{}\renewcommand{\RefusalBestCandidateCost}{44.9\%}
\providecommand{\PredictorPairs}{}\renewcommand{\PredictorPairs}{131}
\providecommand{\PredictorBodies}{}\renewcommand{\PredictorBodies}{8}
\providecommand{\PredictorTreeBalancedAccuracy}{}\renewcommand{\PredictorTreeBalancedAccuracy}{0.63}
\providecommand{\PredictorTreeMacroAverage}{}\renewcommand{\PredictorTreeMacroAverage}{0.56}
\providecommand{\PredictorTreeFoldRange}{}\renewcommand{\PredictorTreeFoldRange}{0.43--0.65}
\providecommand{\PredictorTreeFoldsScored}{}\renewcommand{\PredictorTreeFoldsScored}{6}
\providecommand{\PredictorMajorityMacroAverage}{}\renewcommand{\PredictorMajorityMacroAverage}{0.50}
\providecommand{\MatrixPossiblePairs}{}\renewcommand{\MatrixPossiblePairs}{336}
\providecommand{\MatrixOfferedPairs}{}\renewcommand{\MatrixOfferedPairs}{157}
\providecommand{\MatrixFilledCells}{}\renewcommand{\MatrixFilledCells}{136}
\providecommand{\MatrixFill}{}\renewcommand{\MatrixFill}{40.5\%}
\providecommand{\MatrixExtraSolves}{}\renewcommand{\MatrixExtraSolves}{21}
\providecommand{\ExtraSolvesHandCopy}{}\renewcommand{\ExtraSolvesHandCopy}{7}
\providecommand{\ExtraSolvesObjectRepeat}{}\renewcommand{\ExtraSolvesObjectRepeat}{14}
\providecommand{\ExtraSolvesObjectRepeatPass}{}\renewcommand{\ExtraSolvesObjectRepeatPass}{3}
\providecommand{\BodyCartVariantPairs}{}\renewcommand{\BodyCartVariantPairs}{4}
\providecommand{\StatementsTested}{}\renewcommand{\StatementsTested}{5}
\providecommand{\StatementsSurvivingLeaveOneBodyOut}{}\renewcommand{\StatementsSurvivingLeaveOneBodyOut}{3}
\providecommand{\NeverStartEpisodesProbed}{}\renewcommand{\NeverStartEpisodesProbed}{38}
\providecommand{\NeverStartEpisodes}{}\renewcommand{\NeverStartEpisodes}{28}
\providecommand{\PriorRsrdTrials}{}\renewcommand{\PriorRsrdTrials}{90}
\providecommand{\PriorPhysvtwoaTrials}{}\renewcommand{\PriorPhysvtwoaTrials}{80}
\providecommand{\PriorPhysvtwoaSuccess}{}\renewcommand{\PriorPhysvtwoaSuccess}{88.75\%}
\providecommand{\PriorPsiEpisodes}{}\renewcommand{\PriorPsiEpisodes}{20}
\providecommand{\PriorEgoTwoRobotTeleop}{}\renewcommand{\PriorEgoTwoRobotTeleop}{20}
\providecommand{\PriorEgoTwoRobotMorphologies}{}\renewcommand{\PriorEgoTwoRobotMorphologies}{15}
\providecommand{\PriorRtworTrials}{}\renewcommand{\PriorRtworTrials}{1\,050}
\providecommand{\PriorVtwostworTrials}{}\renewcommand{\PriorVtwostworTrials}{10}
\providecommand{\PriorDroidInstitutions}{}\renewcommand{\PriorDroidInstitutions}{13}

\newcommand{\tier}[1]{{\footnotesize\itshape[#1]}}
\newcommand{\SimTier}{\tier{\TierLine}}
\newcommand{\HwTierLine}{kinematic replay $\cdot$ SO-101 $\cdot$ as measured}
\newcommand{\HwTier}{\tier{\HwTierLine}}

\newcommand{\keepwithnext}{\needspace{3\baselineskip}}
\newcommand{\pairs}{(robot, episode)}
\renewcommand{\arraystretch}{1.08}

\title{\vspace{-1.2em}\bfseries Human Demonstrations as Post-Training Data\\
for Any Robot Body:\\ Retargeting the Object, Not the Hand\vspace{-0.4em}}

\author{Nagasrikanth Kallakuri\\ \normalsize Dxtr AI Co.\\
\normalsize\texttt{srikanth@dxtr.co}%
\thanks{The author is the founder of Dxtr AI Co., where this work was done.}
}
\date{}
\begin{document}
\maketitle

\begin{abstract}
Human demonstrations are today used largely for pretraining
vision-language-action models, with little ability to use the same data for
target-embodiment or task-oriented training, which scalable deployment
of robots in industrial and general environments depends on. A
human demonstration recorded on a simple and scalable device such as a
mobile phone can serve as post-training data for a robot whose body differs
in structure and capability from the human who recorded it. This is made
possible not by retargeting the hand's motion but the object's. A capture
yields a body-free record: the object's pose over time, its contact
intervals, and a reconstructed 3D object with provenance. Any robot, given as a URDF and a %
short descriptor, is solved against that record and scored in simulation
against the demonstrated object motion as ground truth: the
human-to-simulation leg of human-to-simulation-to-robot transfer, measured
per robot. Each (robot, episode) pair gets one outcome: pass, failure with a
named mode, or refusal, with its reason, before any motion. The method runs
as a working system over a growing collection; the set analysed here is
a sample of it:
\CorpusCaptures{} captures solved on \CorpusBodies{} robots,
\CorpusPairs{} pairs, of which \OutcomePassShare{} pass,
\OutcomeFailShare{} fail, \OutcomeRefusedShare{} are refused and
\OutcomeUnauditableShare{} are unauditable (\TierLine). Refusals held up: every pair the checks
refuse on stored evidence yet ran anyway failed
(\RefusalPrecisionKNum{} of \RefusalPrecisionKDen{};
\RefusalPrecisionNoClosingJointWeldedNum{} on a body whose public model welds
its hand). A description file could not be shown to screen a robot: its
outcomes were barely more predictable than guessing. No policy has yet
been trained on the record.

The record decides what a robot may post-train on; whether the policy then
works on the real robot is the simulation-to-robot question, answered
downstream.
\end{abstract}

\begin{figure}[t]
  \centering
  \includegraphics{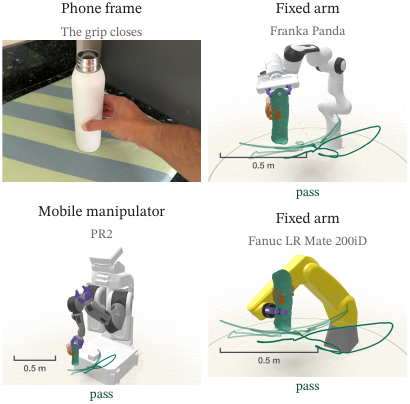}
  \caption{\textbf{One phone capture, multiple robots, multiple passes.} The
  demonstrator's own frame as the grip closes in a pouring episode, and
  the same moment replayed in simulation by two fixed arms and a mobile %
  manipulator, each with its own grasp; orange marks where the hand held
  the object. Nothing is retouched. \SimTier}
  \label{fig:teaser}
\end{figure}

\begin{figure*}[t]
  \centering
  \includegraphics{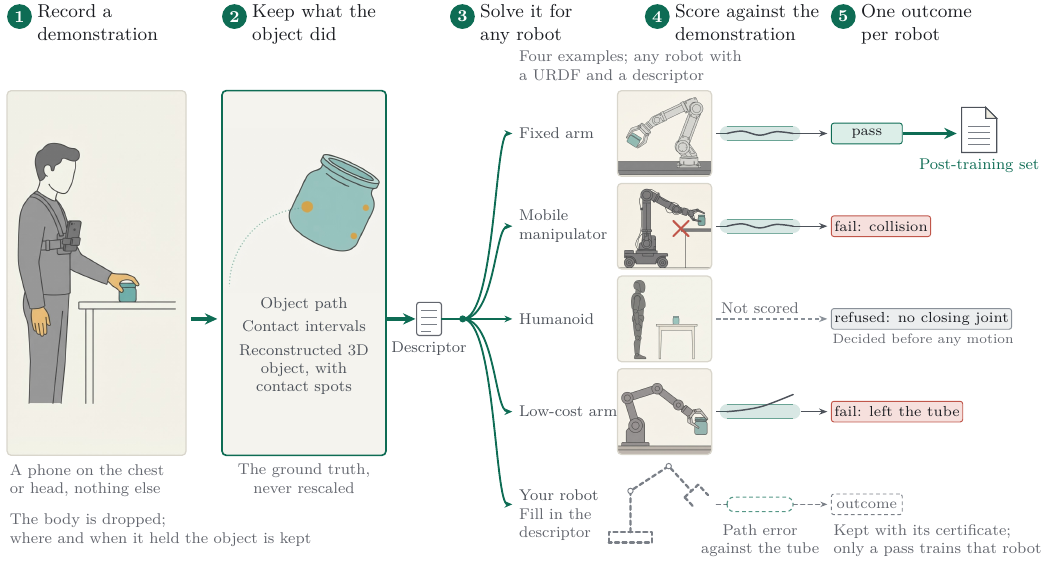}
  \caption{\textbf{One demonstration, one body-free record, an
  answer for every robot.} A demonstration is recorded on a phone worn on
  the body, here the chest. Only what the object did is kept, its path, its contact
  intervals and its reconstructed 3D object, and that is the ground truth. Any robot given as %
  a URDF and a short descriptor is solved against it, and its replay is
  scored against the demonstrated motion by object-path error inside the
  tube. Only a pass enters that robot's post-training set, and a refusal
  is decided before any motion. The bodies
  and the outcomes shown are examples: each outcome is one that a body of
  that class received in \S\ref{sec:corpus}, and a robot the method has
  never seen is admitted by filling in the descriptor. \SimTier}
  \label{fig:concept}
\end{figure*}

\section{Introduction}
\label{sec:intro}

\begin{figure}[tb]
  \centering
  \includegraphics{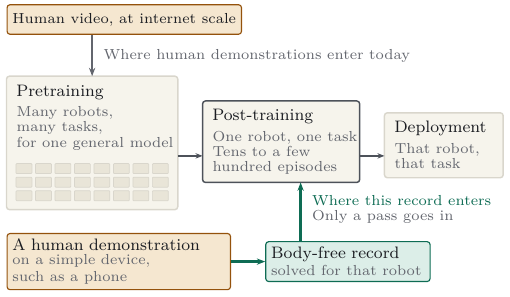}
  \caption{\textbf{Where a demonstration enters.} Human video reaches robot
  learning mostly at pretraining. The record lets one demonstration enter at
  post-training, for one robot and one task, which is where the body decides
  what can be used.}
  \label{fig:enters}
\end{figure}

A human demonstration records what an object did, and a great deal about
the body that did it. Any robot can use the first, but only a body built
like the demonstrator's can use the second. Today, robot learning uses
human demonstrations mostly for
pretraining, where human video at internet scale is retargeted in bulk to
teach a vision-language-action model what manipulation looks
like~\citep{luo2026beingh05,wang2026ego2robot}.
Deployment depends on the step after that (Fig.~\ref{fig:enters}), in
which a specific robot is
post-trained on a specific task from tens to a few hundred episodes of its
own body. Human demonstrations are hard to use there, because the robot's
body decides which of them it can execute at all. Systems that do transfer
a demonstration to a robot target one robot at a
time~\citep{li2024okami,huang2026physv2a,zhai2026psi,yu2025real2render2real},
and they drop the demonstrations that robot cannot use.

This paper turns a demonstration recorded on a simple and scalable device,
such as a mobile phone with nothing attached to it, into post-training data
for any robot (Fig.~\ref{fig:teaser}). It does so by retargeting the object's motion. Throughout, the data is post-training data: a target-embodiment
training set, the tens to a few hundred episodes of one robot on one task on
which a pretrained policy is post-trained, or at that scale a task policy is
trained from scratch. Nothing here is pretraining data. We write
\emph{post-training set} for short.

Fig.~\ref{fig:concept} shows the design, and Table~\ref{tab:glossary}
collects the terms. A capture is reduced to a body-free record: the
object's pose over time in metres, the intervals over which it was held,
and one reconstructed 3D object whose observed surface is kept apart from %
the completed one.
The hand is treated as a witness to contact rather than copied: when and
where it held the object are kept on the object, as the contact intervals
and as a contact map on the mesh. Any robot, given as a
URDF and the short descriptor of Table~\ref{tab:descriptor}, is solved
against that record without rescaling it. Its kinematic replay is then
scored against the demonstrated object motion as ground truth. This
measures, for each robot, the human-to-simulation leg of
human-to-simulation-to-robot transfer (\S\ref{sec:why}). Each (robot,
episode) pair receives one outcome. The pair passes at the declared
tolerance or fails with a named mode, or the robot refuses the episode
before any motion is generated. A refusal carries a reason from a fixed
list and the measured quantity behind it. Every outcome goes to an
append-only ledger in which the latest entry for a run supersedes the
earlier ones. A pass that cannot meet a later check is therefore retracted
by a dated entry, and the earlier row stays in the ledger.

Every pipeline we know of that retargets human demonstrations already makes the
refusal decision. A feasibility filter drops motions a humanoid cannot
track~\citep{he2024h2o,ji2024exbody2}. A grasp planner discards a
trajectory when none of its sampled grasps survives a simulator
check~\citep{zhai2026psi}, and a curation stage removes
whatever a quality model dislikes~\citep{wang2026ego2robot,ye2025video2policy}.
The decision stays inside the pipeline, and only what survives it is
released. A reader then cannot tell whether an episode is missing because
it was never captured, because one body could not reach it, or because a
filter that no longer exists once rejected it. We keep the decision, with
its reason, beside the record.

The set analysed here holds \CorpusCaptures{} captures solved on
\CorpusBodies{} robots from \CorpusBodyClasses{} body classes, \CorpusPairs{} (robot, episode)
pairs in all, of which \OutcomePassShare{} pass, \OutcomeFailShare{} fail,
\OutcomeRefusedShare{} are refused and \OutcomeUnauditableShare{} are
unauditable. Every refused pair that
ran anyway failed (\RefusalPrecisionKNum{} of
\RefusalPrecisionKDen{}, \S\ref{sec:refusalvalidity}). A robot's
description predicts its outcomes barely better than guessing does
(\S\ref{sec:predictor}). Ranking grasps by sweep coverage passed
\AblationGraspRankingPassControl{} of pairs where ranking by grasp quality
passed \AblationGraspRankingPassArm{} (\S\ref{sec:ablations}).

Human2Sim2Robot~\citep{lum2025human2sim2robot} names the human-to-robot %
embodiment gap. It and Video2Sim2Real~\citep{han2026video2sim2real} %
cross that gap for one robot. They use the demonstrated
object motion as a reward for reinforcement learning in simulation, then
transfer the policy. We use the same object motion as ground truth, not as a
reward. We measure the human-to-simulation leg for every robot,
kinematically and with no learning in the loop, and we keep the
measurement. The simulation-to-robot leg belongs to the policy post-trained
on the result and is measured downstream.

\begin{table}[t]
  \centering\scriptsize
  \setlength{\tabcolsep}{3pt}
  \caption{\textbf{Terms used throughout.}}
  \label{tab:glossary}
  \begin{tabular}{@{}p{1.7cm}p{6.0cm}@{}}
    \toprule
    post-training set & a target-embodiment training set: tens to a few
      hundred episodes of one robot on one task \\
    record & the body-free part of a demonstration: the object's pose over
      time, its contact intervals, and its reconstructed 3D object \\ %
    reference motion & the demonstrated object motion, the ground truth every
      robot is scored against \\
    pair & one robot's answer on one solve run; a capture solved under more
      than one retargeting mode contributes one pair per solve run \\
    outcome & a pair's answer at the stated tier: pass, failure with a named
      mode, refusal with a reason, or unauditable \\
    tube & the declared tolerance around the reference motion,
      \ConstTubePos{} and \ConstTubeOri \\
    tube coverage & the fraction of scored frames inside the tube; a pass
      needs \ConstPassCoverage \\
    human-to-sim gap (the human-to-simulation leg) & the object-path error
      of a robot's kinematic replay
      against the reference motion; tube coverage is its summary \\
    tier line & the claim a number may support, printed with it; a
      simulated outcome and a hardware measurement never share one \\
    witness & the evidence certifying a frame as contact: a pinch bar and a
      fingertip trigger together certify touch; object co-motion may extend
      a touch-certified span and never open one \\
    headroom & a body property over the demonstration's demand: reach over
      path extent, usable stroke over object width \\
    \bottomrule
  \end{tabular}
\end{table}

\keepwithnext
\paragraph{Contributions.}
\begin{enumerate}\setlength{\itemsep}{1pt}\setlength{\parskip}{0pt}\setlength{\topsep}{2pt}
\item The body-free demonstration record, and the per-robot
  human-to-simulation gap it makes measurable (Fig.~\ref{fig:concept},
  Fig.~\ref{fig:record}, \S\ref{sec:record}).
\item Refusal before motion, from a fixed vocabulary and kept with its
  certificate. Refused pairs that ran anyway all failed, and we could not screen
  a robot the model has not seen from its description
  (\S\ref{sec:refusalvalidity}, \S\ref{sec:predictor}).
\item Grasp selection ranked by sweep coverage of the whole demonstrated path,
  with its ablation against ranking by grasp quality
  (Alg.~\ref{alg:grasp}, Table~\ref{tab:ablations}).
\item A worked corpus across \CorpusBodies{} embodiments in
  \CorpusBodyClasses{} classes, with retractions on the record
  (Fig.~\ref{fig:outcomes}, Table~\ref{tab:bodies},
  Table~\ref{tab:retractions}).
\end{enumerate}

\section{The object as ground truth}
\label{sec:why}

\begin{figure}[tb]
  \centering
  \includegraphics{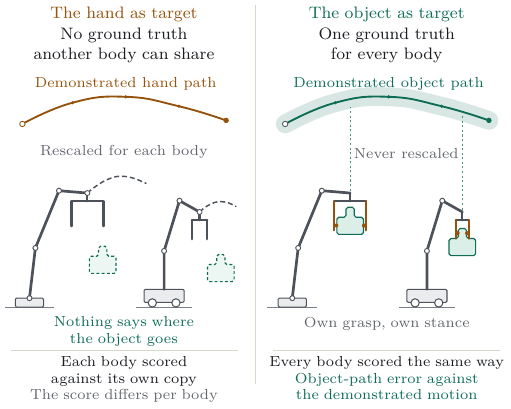}
  \caption{\textbf{The hand as target, the object as target.} Copying the
  hand gives a body of a different build nothing to be measured against.
  Retargeting the object gives every body the same target, scored the same
  way.}
  \label{fig:handobject}
\end{figure}

Copying the hand has no ground truth once the body changes. A hand path
scaled onto the jaw gripper of a low-cost arm is a guess about what that
arm should do, and the guess differs for every body. Robots given the same
demonstration then cannot be compared with each other, and none of them
can be said to have reproduced it (Fig.~\ref{fig:handobject}). The object's
motion is different: it is what the task is about, it is the same for
every body, and a robot either reproduces it within a stated tolerance or
does not. We therefore take the ground truth from the object: its pose over time
and the intervals over which it was held. Both are recovered once from the
capture, in metres, and never rescaled to suit a workspace.

This makes outcomes comparable across bodies, since every body is scored
against the same reference motion with the same tolerance. It also makes
the gap between the demonstration and the simulated robot a per-frame
quantity, the object-path error, summarised by the fraction of scored
frames inside the tube (\S\ref{sec:solve}). Some episodes can be ruled out
from a robot's description and the record alone: its end effector has no
closing joint, its stroke is narrower than the object, or its reach is
shorter than the sweep. An attempt that cannot pass spends solve time and, worse, would
put a motion the body cannot execute into its post-training set. So the
robot refuses before any motion, and the measured quantity goes on the
record.

The choice has costs, however. The object must be rigid, with a
recoverable pose, so a deformable object or a hand-to-hand transfer is
refused as object not rigid (Table~\ref{tab:refusals}). Only the
contact interval is scored, because the approach and the
retreat are the least transferable part of a demonstration, and each body
therefore gets its own synthesised approach and retreat. The demonstrated path is a tolerance target for
reproducing the task, not an optimum, and a robot that finds a better path
is scored as having missed this one.

Other parts of a demonstration do not depend on the body either: the scene
the object moved through, the demonstrator's path across it, the state
changes the task caused and the sounds it made. The record in this paper
holds the object's motion, its contact intervals, its reconstructed 3D %
object, and the support
plane and scene bounds that the solver reads for collision and placement.
The demonstrator's path is formulated for bodies that move but has no
gated outcome yet. Environment state and audio are left as extensions of
the same record. Fig.~\ref{fig:record} shows what is stored.

\section{Related work}
\label{sec:related}

\begin{table*}[t]
  \centering\footnotesize
  \caption{\textbf{Where the record differs.} The closest systems and this
  one, as their sources state them. ``Internal''
  means the system decides and does not publish the decision.}
  \label{tab:related}
  \begin{tabular}{@{}>{\raggedright\arraybackslash}p{2.55cm}>{\raggedright\arraybackslash}p{2.6cm}>{\raggedright\arraybackslash}p{2.6cm}>{\raggedright\arraybackslash}p{2.3cm}>{\raggedright\arraybackslash}p{2.85cm}>{\raggedright\arraybackslash}p{3.7cm}@{}}
    \toprule
    Method & Input & Per-robot data & Bodies & Verification &
    Failures / refusals published \\
    \midrule
    OKAMI~\citep{li2024okami} & 1 RGB-D human video & robot rollouts for the
      policy & 1 humanoid & hardware & aggregate failure modes \\
    RSRD~\citep{kerr2024rsrd} & 1 video + object scan & none & 1 bimanual &
      hardware, \PriorRsrdTrials{} trials & per-phase and end-to-end rates \\
    PhysV2A~\citep{huang2026physv2a} & video 6-DoF object motion + RGB-D &
      kinematic model & 1 arm & hardware, \PriorPhysvtwoaTrials{} trials,
      \PriorPhysvtwoaSuccess{} & gates grasp--path pairs, then ranks the
      survivors continuously; decision not published \\
    PSI~\citep{zhai2026psi} & RGB-D human videos per task & none (sim as a
      filter) & 1 arm & hardware, \PriorPsiEpisodes{} episodes per task &
      keeps a per-pair feasibility label and trains a scorer on it; not
      published as a record \\
    Human2Sim2Robot \citep{lum2025human2sim2robot} & 1 RGB-D human video &
      none; RL in simulation per task & 1 arm + dex hand & hardware &
      per-task success rates only \\
    Video2Sim2Real~\citep{han2026video2sim2real} & 1 video, fixed RGB-D & none &
      1 arm + dex hand & hardware, \PriorVtwostworTrials{} trials per task &
      per-task misses \\
    Ego2Robot~\citep{wang2026ego2robot} & egocentric video &
      \PriorEgoTwoRobotTeleop{} teleop demos per task &
      \PriorEgoTwoRobotMorphologies{} synth., 1 real & hardware, 5 tasks &
      internal curation \\
    EgoInfinity~\citep{wang2026egoinfinity} & internet RGB video & none &
      G1, Robonaut2, two FR3 (retargeting) & retargeting metrics; no task success reported & qualitative only \\ %
    R2R2R~\citep{yu2025real2render2real} & phone scan + 1 video & none &
      1 bimanual & hardware, \PriorRtworTrials{} trials & qualitative
      modes, on a second arm only \\
    DROID~\citep{khazatsky2024droid} & teleoperation & yes, by definition &
      \PriorDroidInstitutions{} institutions, one shared platform & none &
      unsuccessful episodes released, one body \\
    \midrule
    \textbf{This work} & one phone, nothing attached & none; a URDF and a
      descriptor & \CorpusBodies{} bodies, \CorpusBodyClasses{} classes &
      kinematic replay against a declared tube & \textbf{per robot: pass,
      named failure or pre-flight refusal from a closed vocabulary, with
      retractions} \\
    \bottomrule
  \end{tabular}
\end{table*}

\paragraph{Object-centric transfer from human video.}
Transferring the object's motion instead of the hand's is an established
line of work. Single-video systems replay a recovered object
trajectory~\citep{li2024okami,heppert2024ditto,kerr2024rsrd,zhu2024orion},
and a policy can be trained to generate the object's pose trajectory from
a few demonstrations~\citep{hsu2024spot}.
Flow and track methods transfer point
motion~\citep{wen2024atm,bharadhwaj2024track2act,xu2024im2flow2act,yuan2024generalflow,chen2025ecflow},
and real-to-sim-to-real systems use the demonstrated object motion as a
reward~\citep{dan2025xsim,lum2025human2sim2robot,han2026video2sim2real}.
Capture on a bare phone is not new
either~\citep{palanisamy2026mobileego},
nor is retargeting a demonstration to a body given only as a
URDF~\citep{yin2026feasiblecap,wang2026ego2robot};
cross-embodiment policy transfer can also mask the end effector out of
the policy's view instead of retargeting~\citep{piseno2026cloak}.
A handheld gripper, with or without a phone on
it~\citep{chi2024umi,etukuru2024rum,xu2026hommi,patel2026bpp,fang2026robopocket}, or a
motion-capture glove~\citep{wang2024dexcap} records the end effector
itself and so bypasses retargeting at the price of the device; a phone with nothing
attached records the object and leaves the end effector to the
descriptor. MimicGen and DexMimicGen~\citep{mandlekar2023mimicgen,jiang2024dexmimicgen}
also cut demonstrations into object-centric segments, to generate robot
data in simulation from a few robot demonstrations, and EgoMimic~\citep{kareer2024egomimic}
co-trains on egocentric human video and robot data for one platform; the
record keeps the segments as a measurement per body instead. The record
itself is recovered with off-the-shelf components, named in
Appendix~C~\citep{lugaresi2019mediapipe,pavlakos2024hamer,ravi2024sam2,xiang2025trellis}.
Table~\ref{tab:related} lists no system that releases, per robot, the
episodes it could not use and why.

\paragraph{Grasp selection conditioned on the whole motion.}
Choosing a grasp because the motion after it stays feasible is an old
idea. The grasp is a planning variable in~\citet{simeon2004manipulation},
filtered by reachability in~\citet{xue2011reachability}, chosen by
post-grasp objectives in~\citet{ghalamzan2016taskrelevant,ghalamzan2018postgrasp},
optimised with the trajectory by~\citet{wang2020manipulationtrajopt} and
ranked under predicted object motion by~\citet{akinola2021dynamicgrasping}.
\citet{huang2026physv2a} rank the grasp--path pairs that pass their
gate, and \citet{zhai2026psi} discard a trajectory when every sampled
grasp fails in simulation, then train a scorer on that label. See
also~\citet{trupin2025itup,liu2026executionaware,shi2026howtransfer}.
We claim none of this for Alg.~\ref{alg:grasp}, whose base search follows
reachability and placement
work~\citep{zacharias2007capability,vahrenkamp2013placement,porges2014reachability}.
What we add is to keep, per robot, the coverage and the refusal the
algorithm produces.

\paragraph{What happens to the discards.}
Feasibility filtering before training is standard, in humanoid motion
retargeting~\citep{he2024h2o,ji2024exbody2}, at capture
time~\citep{yin2026feasiblecap}, by command-following checks over generated
videos~\citep{patel2026rigvid} and by success in
simulation~\citep{ye2025video2policy}. None keeps the discard as a
per-robot outcome with a reason. DROID does release the episodes its
collectors marked unsuccessful, on one platform~\citep{khazatsky2024droid}.
A benchmark on the SO-101 arm classifies rollout failures after
execution~\citep{yu2026so101bench}, and Dex-Net predicts a grasp's
probability of success~\citep{mahler2017dexnet}. Our refusal is a geometric decision,
made before anything moves.

\section{The record}
\label{sec:record}

\begin{figure*}[t]
  \centering
  \includegraphics{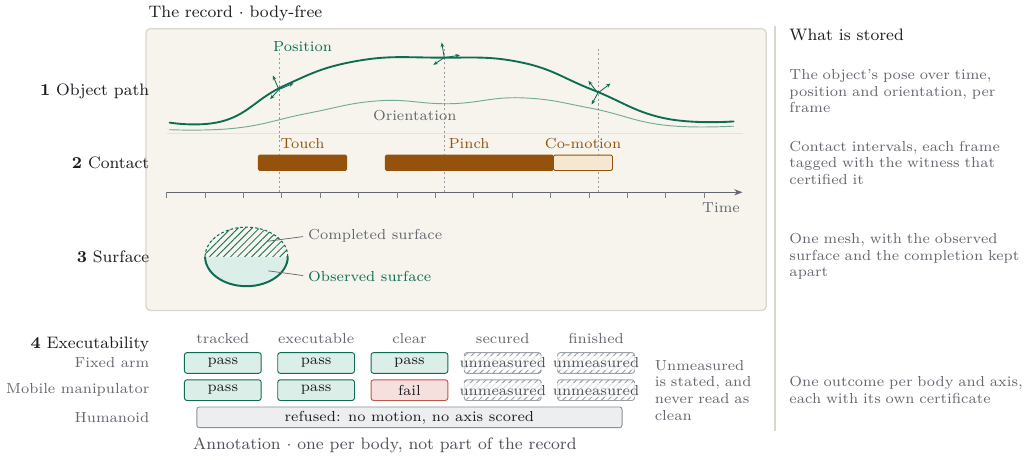}
  \caption{\textbf{What an episode carries, and what a body adds to it.}
  The record (left, cream) is body-free. It holds the object's pose over
  time, the contact intervals with the witness that certified each frame,
  and one mesh that keeps the observed surface apart from the completion.
  The executability block (bottom) is an annotation with one row per body,
  and it never edits the record. An axis nobody measured is hatched and is
  not read as a pass. Only a pass enters a body's post-training set. A named
  failure or a refusal is kept with its certificate and stays out. \SimTier}
  \label{fig:record}
\end{figure*}

The record is what a capture keeps once the demonstrator's body is taken
out. A robot's outcome is stored beside it and never edits it, so adding
a robot cannot change an earlier robot's outcome (Fig.~\ref{fig:record}).

\paragraph{The reference motion.} A capture yields the object's pose over
time, one pose per frame on a uniform \ConstFrameRate{} clock, with
position in metres and orientation as a unit quaternion. The poses are
recovered by registering masked depth clouds to a fused reference cloud.
The capture also yields the intervals over which the object was held, each
frame tagged with the witness that certified it. Together these form the
reference motion. The commanded opening comes from the object's geometry, and the demonstrated pinch only
times opening and closing. The mesh flags each vertex as observed or
completed, so which surface a chosen grasp touches can be seen. Where the
hand touched the surface is kept on the mesh as a contact map. It biases
the choice among candidate grasps and never filters them (Alg.~\ref{alg:grasp}),
so a robot may grasp where the person did not.

\paragraph{Outcomes.} Each \pairs{} pair gets one outcome: pass, a named
failure from the verifier's list of modes, a refusal from the pre-flight
list, or unauditable when the stored bundle is missing and nothing may be
claimed. The outcome has axes, scored separately: \emph{tracked}, the
object pose inside the tube over the scored frames; \emph{executable},
the motion-quality gate; \emph{clear}, no interpenetration of the scene,
the body's own links or the object with a non-jaw link; \emph{secured},
the grasp-closure checks; and \emph{finished}, the phase decomposition
through release. A pass at the kinematic tier (tier~K in the figures) needs
tracked, executable and clear; secured and finished belong to higher
tiers, quoted once in \S\ref{sec:refusalvalidity} from the supplement's
re-score ledger. An axis nobody measured is recorded as unmeasured and is
never read as a pass.

\begin{figure}[tb]
  \centering
  \includegraphics{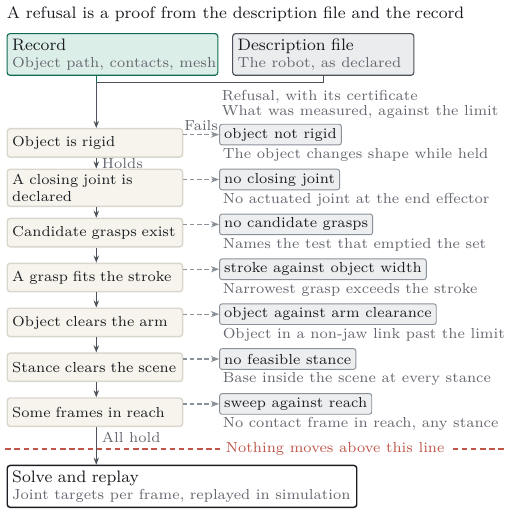}
  \caption{\textbf{Refusal before motion.} The pre-flight checks read the
  description file and the record. The first check that fails ends the
  pair in a refusal with its certificate, and the robot does not move. A
  pair that clears every check is solved and replayed.}
  \label{fig:refusal}
\end{figure}

\paragraph{Refusals.} A refusal names one of \RefusalVocabularySize{}
reasons, each provable at solve time (Fig.~\ref{fig:refusal},
Table~\ref{tab:refusals}). The robot holds its rest pose, and the refusal
is stored with its certificate, which gives the quantity measured and the
limit it missed. Partial coverage is not a refusal. When a grasp covers
only part of the demonstrated interval, the pair is attempted, and the
covered fraction is recorded beside the outcome as a quantity of its own.

\begin{table}[tb]
  \centering\footnotesize
  \setlength{\tabcolsep}{3pt}
  \caption{\textbf{The refusal vocabulary.} Each reason, the quantity its
  certificate carries and the limit that quantity missed, in the order of
  Fig.~\ref{fig:refusal}. No check reads a replay.}
  \label{tab:refusals}
  \renewcommand{\arraystretch}{1.25}
  \begin{tabular}{@{}>{\raggedright\arraybackslash}p{1.95cm}>{\raggedright\arraybackslash}p{3.45cm}>{\raggedright\arraybackslash}p{2.65cm}@{}}
    \toprule
    reason & quantity measured & what it missed \\
    \midrule
    object not rigid & held frames that fit the object's shape at rest &
      too few, on an object declared deformable \\
    no closing joint & actuated joints at the end effector, from the
      descriptor & none declared \\
    no candidate grasps & antipodal pairs in both friction cones, with a
      clear approach & none survive sampling; the test that emptied the
      set is named \\
    stroke against object width & usable stroke against the narrowest
      candidate & every candidate is wider \\
    object against arm clearance & object depth inside a non-jaw link &
      over \ConstClearanceGate{} for every candidate \\
    no feasible stance & a wheeled base's clearance from the scene &
      inside it at every stance \\
    sweep against reach & coverage of the contact frames & zero at every
      searched stance \\
    \bottomrule
  \end{tabular}
\end{table}

\paragraph{The descriptor.} A robot is read only through the descriptor
of Table~\ref{tab:descriptor}. Nothing downstream branches on a robot's
name, so a robot outside the set studied here is admitted by filling in
that table. Supplement~\S S2 gives every field and what happens when it is
left blank.

\begin{table}[tb]
  \centering\footnotesize
  \setlength{\tabcolsep}{3.5pt}
  \caption{\textbf{The body descriptor.} What the method reads out of a
  body description beyond the model's own joint list, joint limits and
  rest pose, what reads it, and what it can decide. A property the
  description omits still produces an answer.}
  \label{tab:descriptor}
  \renewcommand{\arraystretch}{1.2}
  \begin{tabular}{@{}>{\raggedright\arraybackslash}p{1.9cm}>{\raggedright\arraybackslash}p{2.15cm}>{\raggedright\arraybackslash}p{3.0cm}@{}}
    \toprule
    property & read by & decides \\
    \midrule
    closing joints & candidate synthesis & a refusal when none is declared;
      a closure is never faked \\
    usable stroke & width filter & a stroke-against-width refusal when
      no candidate fits \\
    reach radius & coverage and placement & a sweep-against-reach refusal
      when no stance covers the interval \\
    wrist actuation & target construction & orientation commanded, or
      position tracked and orientation reported \\
    base type & placement search & whether the stance may also move
      vertically, or follow a base path \\
    joint count & the solver's null space & how much freedom is left once
      the pose is fixed \\
    frozen link lists & collision queries & which links may touch the
      object and which are a violation \\
    grasp-frame transform & every target & reported as calibrated,
      identity or manual (a bare flange), never assumed \\
    \bottomrule
  \end{tabular}
\end{table}

\paragraph{The ledger.} Outcomes go to an append-only log with one row
per run and robot. A row records the source revision, the tube in force,
the per-axis outcome, the witness counts and the hashes of the artefacts
scored. The latest row supersedes earlier ones, and a superseded pass
stays as a dated entry. Later entries have superseded
\RetractedPairShare{} of the pairs: \RetractionCauseArtifactReminted{}
because the bundle was produced again and re-scored,
\RetractionCausePreconditionMeasured{} because a pre-flight check now
refuses the pair before it moves, and none because the same artefact
failed a stricter check (Table~\ref{tab:retractions}).

\section{Method}
\label{sec:method}

\begin{figure*}[t]
  \centering
  \includegraphics{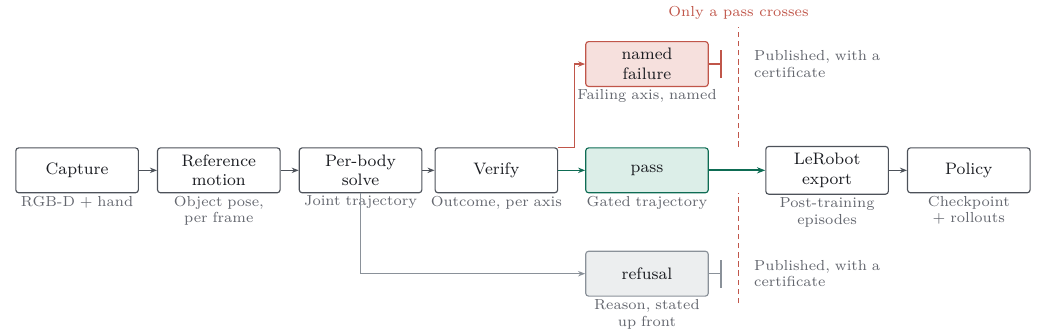}
  \caption{\textbf{Capture to policy, and the two exits.} Only a passing
  trajectory enters the body's post-training set. A named failure and a
  refusal are kept with their certificates and go no further. The refusal is
  decided from solve-time artefacts, before any motion is generated. The
  export and post-training stages are drawn to show where the record sits
  in the chain. This paper's measurements stop at the gate.}
  \label{fig:pipeline}
\end{figure*}

The steps below follow Fig.~\ref{fig:pipeline}, in the order a team would
build them. The grasp choice and the contact vote do the work, and both
read a robot only through its descriptor. Supplement~\S S1 specifies every
stage with its constants.

\subsection{The body description}
\label{sec:bodyfile}

A closure is read from the descriptor as the joints that close and the
opening they can command, so a parallel jaw and a multi-finger hand
enter the same way. A body that declares no closing joint gets no
synthesised grasp and the pair is refused before any motion; its row
records the width it would have needed. Bundles minted before the checks
were enforced instead let the open hand follow the object origin;
\S\ref{sec:refusalvalidity} counts them as ran anyway. A wrist that cannot reach every
orientation tracks position and reports the orientation it achieved. The
freedom it has left is measured by sweeping the free joint, not assumed
from the mechanism. Base type decides how a body may stand. A body bolted
to the work surface may be moved across it but not up or down, and a
wheeled base may go only to stances it can occupy.

\subsection{Grasp selection by sweep coverage}
\label{sec:grasp}

\begin{figure}[tb]
  \centering
  \includegraphics{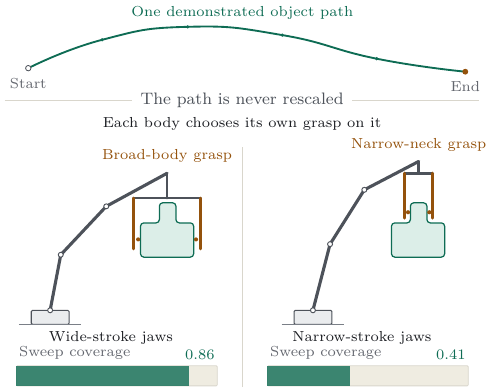}
  \caption{\textbf{One path, two bodies, two grasps.} The demonstrated %
  object path is the same for both arms and is never rescaled. Each body
  picks the grasp that keeps the most of that path solvable under strict
  inverse kinematics, so sweep coverage, not geometric quality, ranks the
  grasps. Conditioning a grasp on the motion that follows is prior
  work~\citep{ghalamzan2016taskrelevant,akinola2021dynamicgrasping,huang2026physv2a,zhai2026psi}.
  The sweep-coverage statistic and the outcome kept per robot are ours.}
  \label{fig:sweep}
\end{figure}

Candidates are generated geometrically, without a learned model
(Alg.~\ref{alg:grasp}). Normals come from the reconstruction, because a
centroid heuristic mislabels the inward faces of non-convex shapes. A
pair of surface samples is a candidate only if the line through its
contacts lies inside both friction cones, which have a half-angle of
\ConstFrictionConeHalfAngle{} at $\mu = \ConstFrictionMu$. A pair outside
them slips under the planner's own friction model, so it is discarded
outright. Candidates are also dropped when the approach
shaft passes through the object or the object would sit more than
\ConstClearanceGate{} inside a non-jaw link.

Sweep coverage ranks the survivors (Fig.~\ref{fig:sweep}). It is the fraction
of contact frames on which a strict inverse-kinematics solve meets both
position and orientation. Ties go to grasp quality. A full solve on a body
with a torso and a second arm costs tens of milliseconds, so the search is
staged: a reach test with no inverse kinematics, then a sparse keyframe
probe, then a dense sweep of the survivors. When a fixed mount cannot cover
the interval, the search moves where the robot stands, over a coarse grid.
The path itself is never rescaled. Coverage does not see collision. The ranking
cannot tell that a grasp it prefers will put the arm through the table or
through itself, and collision is the largest failure class, at
\FailureCollisionShare{} of failures (\S\ref{sec:channels}).

A search that ends without a grasp writes its reason into the grasp block,
and the outcome is read from that field. With no grasp, sweep against
reach means none was found. Beside a chosen grasp, the same certificate
means that some contact frames were left uncovered and the last feasible
pose was held.

\begin{algorithm}[tb]
\footnotesize
\DontPrintSemicolon
\KwIn{canonical mesh $M$ with per-vertex normals and observed/completed
  provenance; contact heat $h$; object path $T_{\mathrm{obj}}(t)$ over the
  demonstrated contact interval $C$; body $B$ with URDF, usable stroke
  $s$, reach radius $r$, grasp-frame transform; constants
  $\mu = \ConstFrictionMu$, cone half-angle \ConstFrictionConeHalfAngle,
  clearance gate \ConstClearanceGate, and the rest of the constants table,
  Supplement~\S S1.15}
\KwOut{grasp $G$, placement $b$, per-frame reachability mask, sweep coverage
  $\kappa$ --- \emph{or} a refusal reason written into the grasp block}
\BlankLine
\lIf{$B$ declares no closing joint}{\Return \textsc{no closing joint}}
$P \leftarrow$ area-weighted surface samples on $M$, fixed seed\;
$A \leftarrow$ antipodal pairs of $P$ whose contact line lies inside both
  friction cones\;
\lIf{$A = \emptyset$}{\Return \textsc{no candidate grasps}, naming the test that emptied it}
$A \leftarrow \{a \in A : \mathrm{width}(a) \le s\}$\;
\lIf{$A = \emptyset$}{\Return \textsc{stroke against object width}}
$A \leftarrow$ poses from $A$: closing axis on the contact line, a fixed set
  of approach rolls, shafts clear of $M$\;
$A \leftarrow \{a \in A :$ the object at $a$ penetrates no non-jaw link of
  $B$ by more than \ConstClearanceGate$\}$\;
\lIf{$A = \emptyset$}{\Return \textsc{object against arm clearance}}
$q \leftarrow$ geometric mean of antipodal margin, stroke margin and local
  flatness, biased (never filtered) by contact heat $h$; deduplicate\;
$\mathcal{B} \leftarrow \{$nominal placement$\}$ extended by a coarse stance
  grid across the surface; \lIf{$B$ is neither bolted to the work surface
  nor on a cart}{extend $\mathcal{B}$ in height as well}
\ForEach{$b \in \mathcal{B}$}{
  \textbf{Stage A} order $A$ by the fraction of $\mathrm{EE}_b(t)$ inside the
    reach annulus of $r$ (no IK)\;
  \textbf{Stage B} probe a few spread keyframes at a reduced iteration
    budget, warm-started; stop early once enough clear the target\;
  \textbf{Stage C} dense strict IK over all $t \in C$ for the top
    survivors; $\kappa \leftarrow |\{t : \text{strict solve}\}| / |C|$\;
}
\lIf{every $b \in \mathcal{B}$ puts $B$'s base inside the scene}{\Return \textsc{no feasible stance}}
$(G,b) \leftarrow \arg\max$ lexicographically over $(\kappa, q)$\;
\lIf{$\kappa = 0$}{\Return \textsc{sweep against reach}}
\Return $(G, b, \text{mask}, \kappa)$; uncovered frames hold the last
  feasible pose and are reported\;
\caption{Grasp selection by sweep coverage, with a staged base search.}
\label{alg:grasp}
\end{algorithm}

\subsection{The contact vote}
\label{sec:contact}

\begin{figure}[tb]
  \centering
  \includegraphics{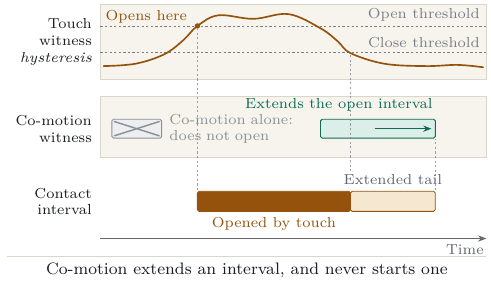}
  \caption{\textbf{Touch opens a contact, and co-motion may only extend it.}
  The object-side witness certifies frames that a fingertip distance
  cannot, as in a raised carry. It is clamped to the touch-only onset, so
  the first frame of an interval is exactly what it was before the witness
  existed. Over the \ContactScoredFrames{} scored contact frames of the
  \ContactPairsWithEvidence{} pairs that carry the witness split,
  \ContactCoMotionShare{} are held by co-motion alone.}
  \label{fig:witness}
\end{figure}

Contact intervals decide which frames are scored and where the grasp must
exist, and \ContactWitnessCount{} witnesses vote on them
(Alg.~\ref{alg:contact}). A pinch bar thresholds the thumb-to-index
distance at a fraction of the episode's own open-to-closed range, since a
fixed threshold does not transfer across hand-pose backends. A fingertip
trigger opens below \ConstContactEnter{} from the surface and closes only
above \ConstContactRelease{}. Every binary threshold is such a hysteresis
pair, because a single threshold at the noise level chatters.

Both read the hand, and both fail on a raised carry, where the fingertip
distance to a partly completed mesh stays large while the object is
plainly held. The third witness reads the object. The palm's offset,
expressed in the object's own frame, stays constant for a rigidly held
object however the object moves. It counts as rigid when it varies by no
more than \ConstCoMotionTol{} over \ConstCoMotionWindow{}
frames, a tolerance set from the capture's own measurement noise.

Co-motion may extend a contact and never open one (Fig.~\ref{fig:witness}).
A co-motion run counts only if it abuts a touch-certified span, and no
frame before its first touch-certified frame is certified. Grasp onset,
which the pre-flight checks key off, is therefore exactly what touch alone
gives. The rule acts on
\NeverStartEpisodes{} of the \NeverStartEpisodesProbed{} episodes whose
stored evidence could be replayed (Supplement~\S S1.6). Its effect on
outcomes was re-scored, not re-solved: with the object witness withheld,
onset is unchanged on every one of the \AblationOnsetHeldDen{} re-scored
episodes (\AblationOnsetHeldNum{} of \AblationOnsetHeldDen{}), the scored
window changes, and no outcome changes (Table~\ref{tab:ablations}). The
denominators differ because the never-start probe replays stored evidence
on \NeverStartEpisodesProbed{} episodes, while the re-score covers
\AblationContactIntervalsEpisodes{} of the ablation set's
\AblationEpisodes{} episodes.

\begin{algorithm}[tb]
\footnotesize
\DontPrintSemicolon
\KwIn{hand landmarks and object poses per frame; reconstructed surface;
  onset \ConstContactEnter, release \ConstContactRelease, dwell
  \ConstContactDwell, co-motion tolerance \ConstCoMotionTol{} over
  \ConstCoMotionWindow{} frames, displacement floor
  \ConstCoMotionDisplaced, all from the constants table,
  Supplement~\S S1.15}
\KwOut{contact intervals; per-frame witness tag in
  \{\textsc{touch}, \textsc{co\_motion}, \textsc{none}\}}
\BlankLine
$T \leftarrow$ frames where the pinch bar is closed (median-filtered
  thumb-to-index distance below a run-relative fraction of this episode's
  own open-to-closed range) \emph{and} a Schmitt trigger on the minimum
  fingertip-to-surface distance is open (below \ConstContactEnter, closing
  above \ConstContactRelease, dwell \ConstContactDwell)\;
\ForEach{frame $t$ with a tracked hand}{
  $o(t) \leftarrow R_{\mathrm{obj}}(t)^{\!\top}(p_{\mathrm{palm}}(t) -
    t_{\mathrm{obj}}(t))$\;
  $\mathrm{rigid}(t) \leftarrow$ worst deviation of $o$ from its window
    mean $\le$ \ConstCoMotionTol{}, and the pose was not propagated from the
    hand\;
  $\mathrm{open}(t) \leftarrow \mathrm{rigid}(t) \wedge$ object measured
    moving\;
  $\mathrm{hold}(t) \leftarrow \mathrm{rigid}(t) \wedge$ (moving $\vee$
    displaced $>$ \ConstCoMotionDisplaced{} from rest)\;
}
\ForEach{maximal run $R$ of \textrm{hold} frames}{
  \lIf{$R$ has no \textrm{open} frame, or abuts no span of $T$}{discard
    $R$}
  \textbf{never-start:} drop every $t \in R$ before the first
    touch-certified frame that $R$ abuts\;
  tag the survivors \textsc{co\_motion} and merge them into $T$\;
}
\Return the merged intervals with their per-frame tags\;
\caption{Contact-interval vote. Co-motion may extend an interval and may
never open one.}
\label{alg:contact}
\end{algorithm}

\subsection{Solving, replay and scoring}
\label{sec:solve}

With a grasp and a stance chosen, each frame gets an end-effector target
tagged with its source: parked, a synthesised approach or retreat, an
observed or interpolated object pose, or a held pose where the grasp
cannot reach. Frames are solved by damped least squares inside a loop
that keeps only the arm chain's own joints and holds the others at a fixed
posture, so the whole-body solver cannot reach a target by moving the
torso or the other arm.

The trajectory is then replayed kinematically against a table plane at
the lowest height the demonstration reached, with the object attached to
the grasping link at its onset pose over the contact interval. This keeps
the tier cheap and also limits what it can claim, since an arm holding
nothing can still score well. Grasp closure is left to a higher tier. The
demonstration and robot worlds are aligned by the solver's own declared
placement translation.

\begin{figure}[tb]
  \centering
  \includegraphics{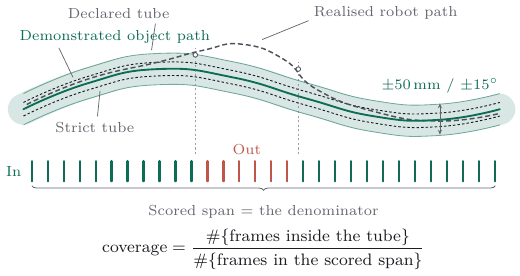}
  \caption{\textbf{The human-to-sim gap, measured.} The demonstrated object
  path is the ground truth and the tube is the declared tolerance.
  Coverage is the fraction of scored frames inside the tube, and a pass
  needs at least \ConstPassCoverage{}. The inner dashed band is the strict
  tube, a tighter reference tolerance recorded beside every outcome as
  strict coverage; the outcome itself is decided on the declared tube.}
  \label{fig:tube}
\end{figure}

\begin{table*}[t]
  \centering\small
  \caption{\textbf{The robots studied.} One row per body, with the
  properties the method reads out of its description and the outcomes it
  was given. The method never reads the first column. One humanoid has a
  three-fingered hand read as a closure with its effective stroke; the
  other wears a five-fingered hand that its published model welds, with no
  finger joint, so it declares no closing joint. \SimTier}
  \label{tab:bodies}
  \setlength{\tabcolsep}{3.5pt}
  \begin{tabular}{@{}llrlrrlrrr@{}}
    \toprule
    body & class & DoF & end effector & stroke (mm) & reach (m) & base &
    pairs & passed & refused \\
    \midrule
    Franka Panda & fixed arm & 7 & parallel jaw & 80 & 0.85 & fixed & 45 & 18 & 5 \\
PR2 & mobile manipulator & 7 & parallel jaw & 80 & 0.82 & mobile & 20 & 8 & 1 \\
myCobot 280 & fixed arm & 6 & parallel jaw & 45 & 0.30 & fixed & 19 & 2 & 3 \\
Unitree G1 & humanoid & 7 & multi-finger hand & 90 & 0.41 & mobile & 19 & 3 & 0 \\
UR5e & fixed arm & 6 & parallel jaw & 85 & 0.85 & fixed & 19 & 7 & 1 \\
Apptronik Apollo & humanoid & 7 & welded hand & -- & 0.53 & mobile & 17 & 0 & 11 \\
Fanuc LR Mate 200iD & fixed arm & 6 & parallel jaw & 85 & 0.72 & fixed & 17 & 11 & 1 \\
SO-101 & low-cost 5-DoF & 5 & parallel jaw & 34 & 0.44 & fixed & 1 & 0 & 0 \\
\crcr
    \bottomrule
  \end{tabular}
\end{table*}

\begin{figure*}[t]
  \centering
  \includegraphics[width=\textwidth]{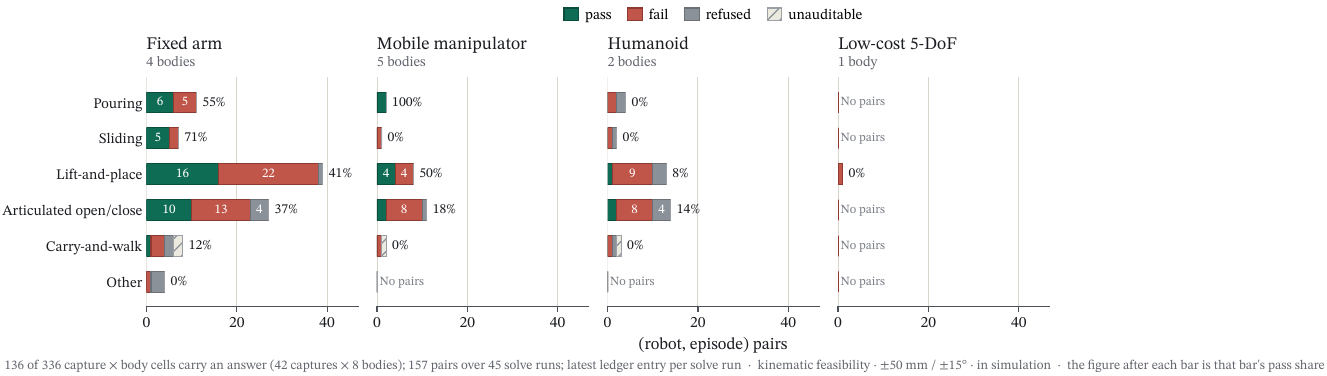}
  \caption{\textbf{Outcomes by task family and body class.} Every pair
  that was offered has an answer here. A body here is a robot on a given base, so an
  arm solved standing on the wheeled cart counts as a mobile manipulator.
  The corpus fills \MatrixFill{} of the capture-by-body matrix; a class
  and family that received no pair is labelled as such rather than drawn
  as a zero. The refusal band shows the pairs that other pipelines
  discard silently. \SimTier}
  \label{fig:outcomes}
\end{figure*}

The human-to-sim gap is measured on this replay. On each scored contact
frame, the gap is the object-path error: the replayed object's distance
from the reference motion, in position and in orientation. Tube coverage
summarises it as the fraction of scored frames inside \ConstTubePos{} and
\ConstTubeOri{} (Fig.~\ref{fig:tube}). A pair passes when at least
\ConstPassCoverage{} of its scored contact frames lie inside the tube and
no failure mode fires. Penetration of the scene, of the robot's
own links or of the object into a non-jaw link fails a pair however well
the object tracked. Supplement~\S S1.13 lists every mode, with the order
in which a failure is named. Coverage is counted over the object-scored
window, which is published with the outcome, and collisions over every
frame the object was held. A frame with no realised object pose counts as
unmeasured, not as clean.

\section{The corpus}
\label{sec:corpus}

The set analysed here is just a small sample of a much larger and growing
collection. It holds \CorpusCaptures{} captures, solved as
\CorpusSolveRuns{} episodes because some captures were solved a second
time under another retargeting mode. Below, \emph{episode} means one
solve. The robots studied are \CorpusBodies{} robots in
\CorpusBodyClasses{} classes, listed in Table~\ref{tab:bodies} with the
properties the method reads: \FleetComposition. Fig.~\ref{fig:teaser}
shows several of them on one episode, and Appendix~B
(Figs.~\ref{fig:strip} to~\ref{fig:appx-grasps}) shows rendered
examples of the phone frame, the record, the replay and the grasp each
body chose. Both humanoids are solved as one arm
on a frozen torso: the torso does not bend, the stance does not shift and
the base does not turn, so their rows measure an arm, not a humanoid. The
low-cost class contributes \BodyLowCostFiveDofPairs{} of the
\CorpusPairs{} pairs, too few to estimate anything. A body here is a robot
on a given base, so an arm solved standing on the wheeled cart counts as a
mobile manipulator, which is why the class panels of
Fig.~\ref{fig:outcomes} count more bodies than Table~\ref{tab:bodies}
lists; \BodyCartVariantPairs{} pairs listed under their arm in
Table~\ref{tab:bodies} are mobile-manipulator pairs there. Of
\MatrixPossiblePairs{} possible capture-by-body cells,
\MatrixFilledCells{} carry an answer, a fill of \MatrixFill{}. The
\MatrixOfferedPairs{} pairs include \MatrixExtraSolves{} repeat solves:
\ExtraSolvesObjectRepeat{} in the object-centric mode
(\ExtraSolvesObjectRepeatPass{} pass) and \ExtraSolvesHandCopy{} under the
hand-copy mode, each a failure, \emph{object could not be scored}, never
a pass (Table~\ref{tab:ablations} withholds its outcome because that
comparison is on geometry alone).
The ledger does not record why the other cells
were not offered, so a class and family with no pair is labelled as such
(Fig.~\ref{fig:outcomes}). Of the \ScopePairs{} pairs with an outcome,
\ScopeContractRows{} carry solve-time evidence and \ScopeScoredWindow{} a
scored window, which is a property of the record and so exists for
refused pairs too. The \OutcomeUnauditable{} unauditable pairs have no
stored bundle, which is why the pass, fail and refusal shares do not add
up to the whole. Every number below is printed with its denominator.

\subsection{Yield per minute of capture}
\label{sec:yield}

\begin{figure}[tb]
  \centering
  \includegraphics[width=\columnwidth]{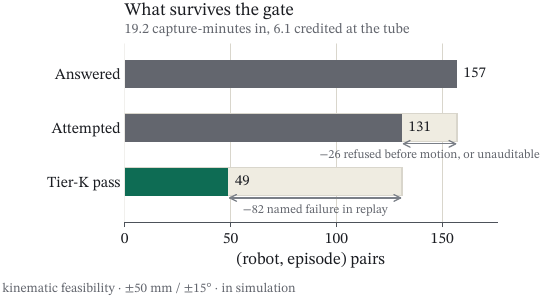}
  \caption{\textbf{Robot-minutes per capture-minute.} Of \CorpusPairs{}
  answered pairs, \OutcomeAttempted{} are attempted and \OutcomePass{}
  pass. The \CorpusCaptureMinutes{} minutes of recorded takes credit
  \CorpusVerifiedRobotMinutes{} robot-minutes at the declared tube. This is
  the cost side of the gate, in the one unit used throughout. \SimTier}
  \label{fig:yield}
\end{figure}

Only part of a recorded take becomes robot motion that passes, and the
record measures that part. The \CorpusCaptureMinutes{} minutes of
recorded takes credit \CorpusVerifiedRobotMinutes{} robot-minutes inside
the declared tube (Fig.~\ref{fig:yield}) \SimTier. A robot-minute is
scored contact time during which the replayed object stayed inside the
tube, credited only to a pair whose replay was executable and clear, so a pair that
collided or never moved the object credits nothing. Approach, retreat and
dead time fill most of a take: the scored window is a median
\ScoredFracTakeMedian{} of the recorded take and \ScoredFracSpanMedian{}
of the demonstrated contact span.

\begin{table*}[t]
  \centering\small
  \caption{\textbf{Ablations.} Each construction is compared with its
  alternative, re-solved on the same pairs and scored the same way (the
  contact vote is re-scored, not re-solved), on \AblationBodies{} fixed
  arms over \AblationEpisodes{} episodes. These runs are kept apart from
  the main ledger. A hand copy earns no outcome in this scoring because it
  does not target the reference motion. Its coverage is measured on the same tube.
  Coverage is the median tube coverage; path ratio is the median length of
  the replayed object path relative to the demonstrated one. \SimTier}
  \label{tab:ablations}
  \begin{tabular}{@{}p{3.4cm}p{5.0cm}rrrr@{}}
    \toprule
    ablation & arm & pairs & pass & median coverage & median path ratio \\
    \midrule
    target: object path vs hand copy & object path & 84 & 28 / 84 (33\%) & 100.0\% & 1.00 \\
 & hand copy (wrist path), scored on the same object tube & 84 & \emph{no outcome} & 18.7\% & 1.56 \\
grasp: coverage-first vs best quality & coverage-first (shipped) & 84 & 28 / 84 (33\%) & 100.0\% & 1.00 \\
 & grasp ranked by geometric quality only (sweep coverage ignored) & 84 & 11 / 84 (13\%) & 85.9\% & 1.13 \\
contact: full witness vote vs touch only & full vote & 55 & 25 / 55 (45\%) & 100.0\% & -- \\
 & touch only & 55 & 25 / 55 (45\%) & 100.0\% & -- \\
stance: base search vs fixed default & base search on (shipped) & 84 & 28 / 84 (33\%) & 100.0\% & 1.00 \\
 & fixed default stance (no base/mount translation search, no rung-1 re-park) & 84 & 26 / 84 (31\%) & 100.0\% & 1.00 \\
\crcr
    \bottomrule
  \end{tabular}
\end{table*}

\subsection{Grasp ranking: sweep coverage against grasp quality}
\label{sec:ablations}

When the same pairs are re-solved, ranking grasps by sweep coverage passes
\AblationGraspRankingPassControl{} of them, against
\AblationGraspRankingPassArm{} for ranking by geometric quality alone
(Table~\ref{tab:ablations}) \SimTier. Most pairs,
\AblationGraspRankingUnchanged{}, keep their outcome, and almost all of
those that change pass only under sweep-coverage ranking. A hand copy gets
no outcome here, and on the same object tube its median tube coverage is
\AblationTargetSourceCovArm{} against \AblationTargetSourceCovControl{}.
The base search raises the pass share from \AblationPlacementPassArm{} to
\AblationPlacementPassControl{}. The contact vote is a re-score, not a
re-solve: with the object witness withheld,
\AblationContactIntervalsPassArmNum{} of \AblationContactIntervalsPairs{}
pairs pass either way, so it changes which frames are scored and no
outcome.

These ablations ran on \AblationBodies{} fixed arms, which are
among the higher-passing robots studied. No mobile, humanoid or
short-reach body was re-solved. The quality-only variant also had help.
When its grasp could not carry the path, a base-motion fallback fired, and
that fallback itself ranks on sweep coverage. It fired several times more
often than under the control, which flatters the quality-only variant, so
the gap above understates what ranking alone buys.

\subsection{Refusal validity, and where refusals come from}
\label{sec:refusalvalidity}

Every refused pair that was run anyway failed \SimTier. Recomputing the
pre-flight checks over the \RefusalScopeRows{} rows that carry solve-time
evidence refuses \ContractRefusals{} pairs. Of those,
\RefusalEnforcedRestPose{} were stored as a rest-pose hold and cannot test
their own refusal. The other \RefusalPrecisionKDen{} were produced before
the checks were enforced and ran anyway. In this natural experiment,
\RefusalPrecisionKNum{} of \RefusalPrecisionKDen{} failed to pass (Wilson
interval \RefusalPrecisionKCI{}), and none reached grasp closure on the
supplement's re-score ledger. The composition matters.
\RefusalPrecisionNoClosingJointDen{} of the \RefusalPrecisionKDen{} were
refused for declaring no closing joint: \RefusalPrecisionNoClosingJointWeldedNum{}
on the welded-hand humanoid, whose open hand followed the object without
holding it, and \RefusalPrecisionNoClosingJointOtherNum{} on bodies whose
early bundles declared no stroke, which \S\ref{sec:adopt} reads as no
closing joint. The other \RefusalPrecisionWithClosureDen{}, refused on
stroke or on an empty candidate set by bodies that do have a closure,
failed as well (\RefusalPrecisionWithClosureNum{} of
\RefusalPrecisionWithClosureDen{}, Wilson interval
\RefusalPrecisionWithClosureCI{}), on a sample too small to bound the rate
tightly. The denominators, in order: \CorpusPairs{} answered pairs, of
which \RefusalScopeRows{} carry solve-time evidence; the ledger stores
\OutcomeRefused{} refusals, \RefusalEnforcedRestPose{} of them inside
those rows; recomputing the checks over those rows refuses
\ContractRefusals{}, of which \RefusalPrecisionKDen{} ran anyway. The
\RetractionCausePreconditionMeasured{} superseded passes of Appendix~A are
not counterexamples: each refusal is of an artefact produced again, not
of the artefact that ran.

\begin{figure}[tb]
  \centering
  \includegraphics[width=\columnwidth]{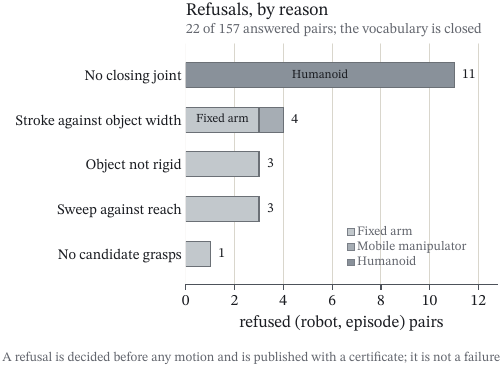}
  \caption{\textbf{Refusals, by reason.} The refusal vocabulary, counted.
  Failure modes among attempts are in the supplement. \SimTier}
  \label{fig:reasons}
\end{figure}

Of the \ContractRefusals{} refusals the checks recompute
(Fig.~\ref{fig:reasons} counts the \OutcomeRefused{} the ledger stores),
\ContractBodyOnlyRefusalShare{} come
from the closure entries of the description file: no closing joint, or
a stroke narrower than every candidate grasp. The one body with no
closing joint, a humanoid whose five-fingered hand is welded in its
published model, accounts for \RefusalNoActuatedGripper{} of the
\OutcomeRefused{} refusals stored in the ledger, \RefusalTopReasonShare{}
of them, so that much of the refused share in the headline is one body's
model. That refusal is about the description, not the hand: the other
humanoid's hand has joints, and is solved. Recall is zero by construction, since
the checks refuse only on proof available at solve time, before any
trajectory is generated, and catch none of
the failures among attempted pairs. Wider refusal boundaries, measured
and not adopted, nest (a negative sweep clearance, then a contact off the
observed surface, then a friction-cone violation) and would have refused
\CandidateExtraRefusedOne{}, \CandidateExtraRefusedTwo{} and
\CandidateExtraRefusedThree{} more pairs. The widest would have caught
\RefusalBestCandidateRecall{} of those failures
(\RefusalBestCandidateRecallNum{} of the \RefusalBestCandidateRecallDen{}
failures with solve-time evidence) while refusing
\RefusalBestCandidateCost{} of the passes, and none would have lost a pair
that the re-score ledger scores as grasp-secured.

\subsection{Screening a new robot from its description}
\label{sec:predictor}
\label{sec:properties}

From the descriptor and the episode's geometry, the best model predicts
the pass or fail label of a robot it has not seen at
\PredictorTreeMacroAverage{} balanced accuracy, averaged over held-out
robots, against \PredictorMajorityMacroAverage{} for always guessing the
majority label \SimTier. The average is over the
\PredictorTreeFoldsScored{} of \PredictorBodies{} robots that have both a
pass and a fail, with a range of \PredictorTreeFoldRange{}. Pooled over
every prediction, the same model scores \PredictorTreeBalancedAccuracy{}.
Pooling rewards a model for recognising which robot a pair belongs to, so
the average over robots is the fairer measure. This does not contradict
\S\ref{sec:refusalvalidity}. The descriptor screens refusals trivially,
since most refusals follow from one of its lines; it is among the
attempted pairs that it could not be shown to predict pass or fail, and
with \PredictorPairs{} attempted pairs over \PredictorBodies{} robots that is a result
we could not show, not one shown impossible. A corpus filtered to its
passes could not be used for this test at all, since it has no negative
label. Grouped by body property, \StatementsSurvivingLeaveOneBodyOut{} of
\StatementsTested{} statements keep the direction of their pass-share gap
when any single robot is left out: reach headroom, base type, and whether
the jaw spans the object (Supplement~\S S7).

\subsection{Refusals and failures, with their denominators}
\label{sec:channels}

Of all answered pairs, \OutcomePassShare{} pass, and of those attempted,
\OutcomePassShareOfAttempted{} \SimTier. The \OutcomeRefused{} refusals use
\RefusalReasonsDistinct{} of the \RefusalVocabularySize{} reasons
(Fig.~\ref{fig:reasons}). Among the \OutcomeFail{} failures,
 collision is the largest class at
\FailureCollisionShare{}, and the largest single mode is \FailureTopMode{}
at \FailureTopModeShare{}.{} Because the object stays
attached to the end effector over the whole scored window,
\texttt{dropped} names a carried path that stays below the demonstrated
one, and \texttt{never\_grasped} an empty scored window.

\section{From the record to a robot}
\label{sec:arm}

A trajectory reaches a robot only as a gated file. The gate runs offline.
It re-parses the description, runs its own forward kinematics and
collision queries, and shares no implementation with the solver it checks.
Its output is a file that carries its own limits and a hash of the
positions, and a file whose positions do not hash to the recorded value is
refused. Position and orientation error are reported but do not gate. The
replay refuses any outcome produced without a measured bench map. The
checks are listed in the supplement. \HwTier

\begin{figure}[tb]
  \centering
  \includegraphics{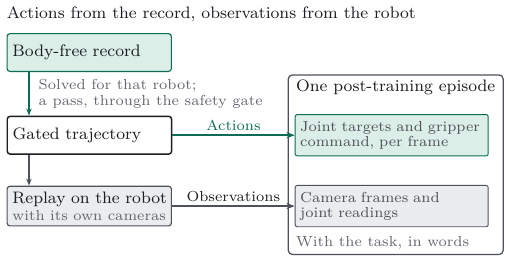}
  \caption{\textbf{Actions from the record, observations from the robot.}
  A passing, gated trajectory supplies the joint targets and the gripper
  command. Replaying it on the robot with its own cameras supplies the
  camera frames and joint readings. With the task described in words,
  these make one post-training episode.}
  \label{fig:episode}
\end{figure}

A post-training episode is also assembled from the gated file
(Fig.~\ref{fig:episode}). The record supplies the action side: the joint
targets and the gripper command for every frame. The observation side,
the camera frames and the joint readings, comes from replaying the gated
trajectory on the target robot with its own cameras. Actions and
observations are then joined, with the task as a string, into one episode
in the LeRobot dataset format.

The record makes an experiment possible, and the supplement fixes its
protocol. Bench tasks are captured with the phone and exported as nested
sets of the same captures: every retargeted trajectory, only those passing
the kinematic tier, and only those also scored on grasp closure. A
teleoperated set on the same tasks and start poses is matched on
robot-minutes. The cheapest part needs no policy. Replaying each
gated episode on the arm and scoring the simulated outcome against what the
arm did tests whether the record predicts anything on hardware.

\paragraph{A pilot on the bench.} We ran that cheapest part on
\PilotTakes{} phone takes of one bench task, \PilotTask{}, carried
through the chain above for the low-cost arm's bench body, its description
as mounted on the bench. Perception
recovered \PilotProcessed{} of them, \PilotTierK{} passed the kinematic
tier \SimTier, \PilotGate{} reached the offline safety gate,

and none was eligible for the replay: every take failed the verifier
(\PilotFailureModes) and the arm stayed parked. The protocol prescribes
five takes on five cells; \PilotTakes{} were filmed, on two of them, and %
the single permitted re-pilot restarts the count. The takes were filmed with
the phone worn on the chest, where the pilot protocol had prescribed a
tripod, and with a thin marker that the hand covered: the
camera saw the marker on at most \PilotVisualCoverageContactMax{} of the
frames in which the hand held it; the rest of its path was propagated from
the hand. The verifier scores reproduction of the demonstrated object path,
so a path the record did not observe cannot pass.

These takes are the protocol's pilot and are not in the corpus of
\S\ref{sec:corpus}; the supplement lists each take and where it stopped.

\section{Adopting the record}
\label{sec:adopt}

To add a robot, supply its kinematic description and the descriptor of
Table~\ref{tab:descriptor}: the closing joints, the usable stroke, reach
radius, wrist actuation, base type, the joint list with limits, the frozen
list of links allowed to touch the object, the grasp-frame transform with
its provenance, and a rest pose. A blank field still yields an answer. A
missing stroke is read as no closing joint, and a missing grasp frame
is derived from the fingertip kinematics, with its provenance recorded. To
add a demonstration, record it on a simple and scalable device such as a
mobile phone, with nothing attached. The mesh is reconstructed from the
episode, so no object scan is needed. Publish, per (robot, episode) pair,
the outcome, its coverage, the window scored, the certificate of any
refusal or failure, and the tier line. Append rows and never edit them.
The latest row supersedes. The tolerances are the same for every body, so
a failure on a robot outside this corpus is described in the same words
as a failure on the robots studied here.

\section{Decisions kept and decisions revised}
\label{sec:generalises}

We would do these again, each within the scope it was measured on:
\begin{itemize}\setlength{\itemsep}{1pt}\setlength{\parskip}{0pt}\setlength{\topsep}{2pt}
\item \textbf{Keep the non-passes.} They made the refusal class testable
  (\S\ref{sec:refusalvalidity}), and they made the screening question
  possible to ask, since a corpus filtered to its passes has no negative
  class (\S\ref{sec:predictor}).
\item \textbf{Rank a grasp by sweep coverage of the demonstrated path.} We
  measured this against ranking by grasp quality on the same pairs, on
  \AblationBodies{} fixed arms (Table~\ref{tab:ablations}).
\item \textbf{Let a witness extend a contact and never open one.} The rule
  acted on \NeverStartEpisodes{} of \NeverStartEpisodesProbed{} episodes.
  Re-scored with the whole witness withheld, no outcome changed; the
  rule's own effect is unmeasured, so its value is in the frames it
  scores, not in the pass count (\S\ref{sec:contact}).
\end{itemize}
We would change these in the next version of the record:
\begin{itemize}\setlength{\itemsep}{1pt}\setlength{\parskip}{0pt}\setlength{\topsep}{2pt}
\item \textbf{Let the ranking statistic see collision.} Sweep coverage
  counts reachable frames and cannot see the arm pass through the table or
  through itself. Collision is the largest failure class, at
  \FailureCollisionShare{} of failures (\S\ref{sec:channels}).
\item \textbf{Refuse below the pass threshold, not only at zero.} A pair
  whose best grasp reaches only part of the path is an
  attempt that cannot pass. It belongs in the refusal class, with its
  sweep coverage as the certificate.
\item \textbf{Model the humanoid's hand.} The one refusal that follows
  from a body's description alone is a five-fingered hand that its
  published model welds. The record never asked for a jaw; the descriptor
  asks for a closing joint, and a hand model with joints answers it the
  way the other humanoid's does.
\item \textbf{Fill the body-by-episode matrix by design.} Captures were
  offered to bodies as they arrived, at a fill of \MatrixFill{}. A planned
  fill would let the property statements of \S\ref{sec:properties} be
  tested. Here they are only observed.
\end{itemize}

\section{Limits}
\label{sec:limits}

Scope, by design.
\begin{itemize}\setlength{\itemsep}{1pt}\setlength{\parskip}{0pt}\setlength{\topsep}{2pt}
\item Outcomes are kinematic \SimTier. A pass says the object's demonstrated
  motion is reproducible by this body at the declared tolerance, with the
  object attached at the grasp. Whether the jaw or hand closed on it, and
  the contact physics, belong to higher tiers.
\item The corpus is a single-demonstrator sample of \CorpusCaptures{}
  captures and \CorpusCaptureMinutes{} minutes of recorded takes. It is
  sized to show the representation and the refusal vocabulary at work, not
  to estimate population rates. Every number regenerates from the ledger.
\item The record assumes a rigid object with a recoverable pose. Deformable
  objects and hand-to-hand transfers are refused as object not rigid.
\item Bodies that carry their own base are solved on a frozen torso. The
  base path is a free variable in the same formulation, solved against the
  object path, and the supplement gives it in full. No ledger entry carries
  a gated base-motion outcome yet.
\item The property groupings are observational and confounded with each
  other. They say which property to screen on, but none is shown to cause
  a pass.
\item Coverage is measured on the scored window, which is a median
  \ScoredFracSpanMedian{} of the demonstrated contact span. Both numbers
  are recorded with every outcome.
\item No policy has yet been post-trained on the record. Before the pilot
  of \S\ref{sec:arm}, one exported trajectory, not a kinematic-tier pass
  and never entered in the ledger, was streamed open-loop to one low-cost
  arm under its safety gate alone with no object present: a bench check of
  the route, not an outcome.{} The pilot of \S\ref{sec:arm} is the first run of the
  hardware protocol, on a bench task outside the corpus; no take reached
  the arm, and the false-pass rate of the \OutcomePass{} corpus passes
  themselves is still unmeasured.{}
  The protocol that answers both is in the supplement.
\end{itemize}

\section{Conclusion}

We took the ground truth for retargeting from the object's motion, the
part of a demonstration that every robot can use. Each body is then scored
against the same reference, and a refusal before motion becomes an answer
kept with its reason. In the sample analysed here, \OutcomeRefusedShare{}
of the answers are refusals, and most of them follow from a line in a
description file. We argue for keeping every answer, retractions included.
The record decides what a robot may post-train on. The next measurement is
what a policy post-trained on it does on the real robot, and the record
lets us make it body by body.

\section*{Disclosure}
The author holds U.S. Provisional Patent Application No. 64/158,349 on the method described here.%

\appendix
\section*{Appendix}
\addcontentsline{toc}{section}{Appendix}

\subsection*{A. Retractions, by cause class}

A retraction is a dated ledger entry that supersedes a published pass.
Later entries have superseded \RetractedPairs{} pairs across
\RetractedCaptures{} captures, \RetractedPairShare{} of the corpus. Most
were withdrawn because the bundle was produced again and re-scored, so the
artefact the pass described no longer exists. The
\RetractionCausePreconditionMeasured{} that a pre-flight check now refuses
were also produced again; the superseding entry is a refusal of the new
artefact, and the earlier artefact's own standing under the checks is not
stored, so these pairs are kept out of the natural experiment of
\S\ref{sec:refusalvalidity} rather than counted on either side. None was
withdrawn because the same artefact failed a stricter check.

\begin{table}[h]
  \centering\small
  \caption{Superseded passes by cause class.}
  \label{tab:retractions}
  \begin{tabular}{@{}lr@{}}
    \toprule
    cause class & superseded pairs \\
    \midrule
    a pre-flight check now refuses the pair & \RetractionCausePreconditionMeasured \\
    bundle produced again and re-scored & \RetractionCauseArtifactReminted \\
    same artefact, stricter check & \RetractionCauseNewVerifierCheck \\
    \bottomrule
  \end{tabular}
\end{table}

\subsection*{B. Rendered examples}
\label{app:renders}

\begin{figure*}[t]
  \centering
  \includegraphics[width=\textwidth]{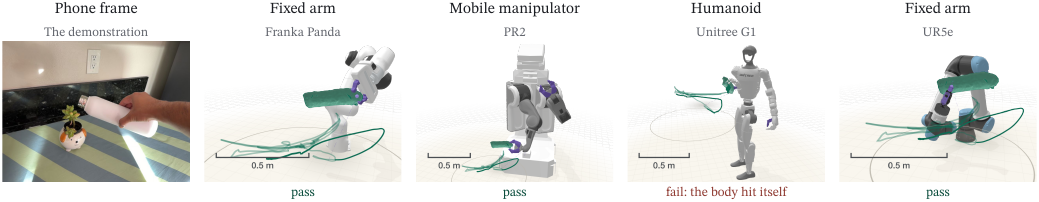}
  \caption{\textbf{One episode, several bodies, each with its own answer.}
  One pouring episode, shown mid-transport, with the
  demonstrated object path in teal; the leftmost tile is the phone's own
  frame at that moment. Each body chose its own grasp and stance
  against the same path, which is never rescaled. Scale bars differ
  because each tile is fitted to its body. Each tile's
  outcome is read from the ledger at build time: a pass, or an attempt
  that failed with its named mode. The thin ring marks the height of the
  work surface. \SimTier}
  \label{fig:strip}
\end{figure*}

\begin{figure*}[t]
  \centering
  \includegraphics[width=\textwidth]{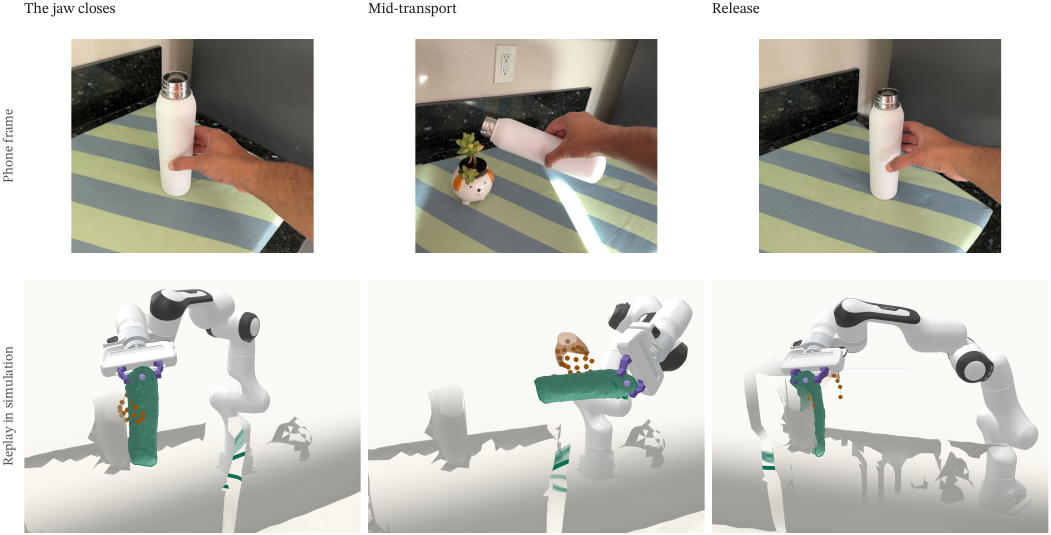}
  \caption{\textbf{From the phone to the replay.} One pouring episode at
  three moments: the demonstrator's grip closing on the object, the carry %
  and tilt over the plant, and the release. Above, the phone's own frame.
  Below, the same moment in the simulated replay of a fixed arm against the
  record: the capture mesh, the reconstructed object on its demonstrated
  path, and the demonstrator's hand as contact marks where it held the
  object. The hand is a witness to contact and is never copied; the arm's
  jaw closes elsewhere on the object, where the path stays reachable for
  that body. This body passed the episode. \SimTier}
  \label{fig:appx-record}
\end{figure*}

\begin{figure*}[t]
  \centering
  \includegraphics[width=\textwidth]{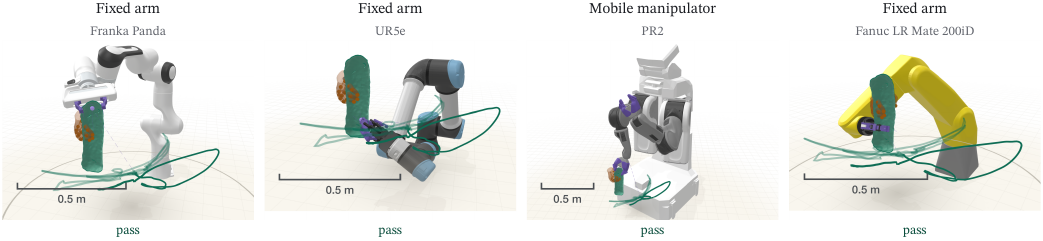}
  \caption{\textbf{The grasp is the body's own.} The same pouring episode
  at the moment the jaw closes, on four bodies that all passed it; the %
  demonstrator's hand at that moment is drawn where it held the object.
  Each body closes its jaw where its own kinematics keep the demonstrated
  path reachable, ranked by sweep coverage (\S\ref{sec:grasp}): near the
  top, low on the body or across the middle. None copies the hand. Each
  tile is fitted to its body, so read the scale bars. \SimTier}
  \label{fig:appx-grasps}
\end{figure*}

Fig.~\ref{fig:teaser} and Figures~\ref{fig:strip}
to~\ref{fig:appx-grasps} are stills of the record and its replays, shot through the viewer's print theme from bundles
in the corpus of \S\ref{sec:corpus}; nothing in them is retouched beyond
that theme's lighting. The phone frames are the captures' own RGB frames
at the moments the replays are shown. In every render the capture mesh is
grey, the reconstructed object is green, the demonstrated object path is
teal, the closing jaw is violet and the demonstrator's hand, where drawn,
is orange. Each tile's body, and the outcome printed under it or stated in
its caption, are read from the ledger when the figure is built, never typed.

\subsection*{C. Reproducibility}

Every number, table and results figure here is regenerated from the
ledger (latest entry per run) by one command, \texttt{make results}, and a
second, \texttt{make check}, fails the build on an unlisted digit typed into the
prose, on a capture identifier and on a page overrun. This version was built on the ledger as of \mbox{\ResultsGeneratedUTC}. A subset of episodes with their per-robot outcomes
is available at \mbox{\url{https://playground.dxtr.co}}; the full ledger, the
re-score ledger and the code are available from the author.

\paragraph{Implementation.} Captures come from the Eye iOS app on an
iPhone with a LiDAR scanner: RGB frames, ARKit scene depth and ARKit
camera poses on one clock. Hands are detected per frame with
MediaPipe~\citep{lugaresi2019mediapipe} and lifted to metric world
coordinates with the LiDAR depth; a second backend regresses an
articulated MANO hand with HaMeR~\citep{pavlakos2024hamer} and registers
it to the LiDAR wrist depth, and the backend used is recorded in each
episode's hands table. The object is segmented by
SAM~2~\citep{ravi2024sam2} from a prompt placed at contact onset, and its %
pose per frame is recovered by trimmed ICP of the masked depth cloud
against a reference cloud fused at onset; a frame whose residual or jump
exceeds its budget takes a pose propagated rigidly from the hand and is
flagged as such. The mesh is reconstructed by the best method available:
TSDF fusion of the masked depth with a symmetry completion is always
computed, and a TRELLIS~\citep{xiang2025trellis} mesh anchored to the
metric cloud replaces it when it clears the anchoring gates; every vertex
carries observed or completed provenance. Inverse kinematics, the
kinematic replay and every collision query run in
PyBullet~\citep{coumans2021pybullet}. The safety gate of \S\ref{sec:arm}
re-implements its forward kinematics and collision checks apart from the
solver.

\bibliographystyle{abbrvnat}
\bibliography{refs_r8}

\begin{thebibliography}{55}
\providecommand{\natexlab}[1]{#1}
\providecommand{\url}[1]{\texttt{#1}}
\expandafter\ifx\csname urlstyle\endcsname\relax
  \providecommand{\doi}[1]{doi: #1}\else
  \providecommand{\doi}{doi: \begingroup \urlstyle{rm}\Url}\fi

\bibitem[Akinola et~al.(2021)Akinola, Xu, Song, and
  Allen]{akinola2021dynamicgrasping}
I.~Akinola, J.~Xu, S.~Song, and P.~K. Allen.
\newblock Dynamic grasping with reachability and motion awareness.
\newblock In \emph{IEEE/RSJ International Conference on Intelligent Robots and
  Systems (IROS)}, 2021.
\newblock arXiv:2103.10562.

\bibitem[Bharadhwaj et~al.(2024)Bharadhwaj, Mottaghi, Gupta, and
  Tulsiani]{bharadhwaj2024track2act}
H.~Bharadhwaj, R.~Mottaghi, A.~Gupta, and S.~Tulsiani.
\newblock {Track2Act}: Predicting point tracks from internet videos enables
  generalizable robot manipulation.
\newblock In \emph{European Conference on Computer Vision (ECCV)}, 2024.
\newblock arXiv:2405.01527.

\bibitem[Chen et~al.(2025)Chen, Li, Huang, Yang, Chen, and
  Wang]{chen2025ecflow}
Y.~Chen, P.~Li, Y.~Huang, J.~Yang, K.~Chen, and L.~Wang.
\newblock {EC-Flow}: Enabling versatile robotic manipulation from
  action-unlabeled videos via embodiment-centric flow.
\newblock In \emph{IEEE/CVF International Conference on Computer Vision
  (ICCV)}, 2025.
\newblock arXiv:2507.06224.

\bibitem[Chi et~al.(2024)Chi, Xu, Pan, Cousineau, Burchfiel, Feng, Tedrake, and
  Song]{chi2024umi}
C.~Chi, Z.~Xu, C.~Pan, E.~Cousineau, B.~Burchfiel, S.~Feng, R.~Tedrake, and
  S.~Song.
\newblock Universal manipulation interface: In-the-wild robot teaching without
  in-the-wild robots.
\newblock In \emph{Robotics: Science and Systems (RSS)}, 2024.

\bibitem[Coumans and Bai(2016--2021)]{coumans2021pybullet}
E.~Coumans and Y.~Bai.
\newblock {PyBullet}, a {Python} module for physics simulation for games,
  robotics and machine learning, 2016--2021.
\newblock \url{http://pybullet.org}.

\bibitem[Dan et~al.(2025)Dan, Kedia, Chao, Duan, Pace, Ma, and
  Choudhury]{dan2025xsim}
P.~Dan, K.~Kedia, A.~Chao, E.~W. Duan, M.~A. Pace, W.-C. Ma, and S.~Choudhury.
\newblock {X-Sim}: Cross-embodiment learning via real-to-sim-to-real.
\newblock In \emph{Conference on Robot Learning (CoRL)}, volume 305 of
  \emph{Proceedings of Machine Learning Research}, pages 816--833, 2025.
\newblock arXiv:2505.07096.

\bibitem[Etukuru et~al.(2025)Etukuru, Naka, Hu, Lee, Mehu, Edsinger, Paxton,
  Chintala, Pinto, and Shafiullah]{etukuru2024rum}
H.~Etukuru, N.~Naka, Z.~Hu, S.~Lee, J.~Mehu, A.~Edsinger, C.~Paxton,
  S.~Chintala, L.~Pinto, and N.~M.~M. Shafiullah.
\newblock Robot utility models: General policies for zero-shot deployment in
  new environments.
\newblock In \emph{2025 IEEE International Conference on Robotics and
  Automation (ICRA)}, pages 8275--8283, 2025.
\newblock \doi{10.1109/ICRA55743.2025.11127857}.

\bibitem[Fang et~al.(2026)Fang, Chen, Xue, Zhou, Le, Wang, Zhang, Lv, Wen, and
  Lu]{fang2026robopocket}
J.~Fang, W.~Chen, H.~Xue, F.~Zhou, T.~Le, Y.~Wang, Y.~Zhang, J.~Lv, C.~Wen, and
  C.~Lu.
\newblock {RoboPocket}: Improve robot policies instantly with your phone, 2026.
\newblock arXiv:2603.05504.

\bibitem[Ghalamzan~E. et~al.(2016)Ghalamzan~E., Mavrakis, Kopicki, Stolkin, and
  Leonardis]{ghalamzan2016taskrelevant}
A.~M. Ghalamzan~E., N.~Mavrakis, M.~Kopicki, R.~Stolkin, and A.~Leonardis.
\newblock Task-relevant grasp selection: A joint solution to planning grasps
  and manipulative motion trajectories.
\newblock In \emph{IEEE/RSJ International Conference on Intelligent Robots and
  Systems (IROS)}, pages 907--914, 2016.
\newblock \doi{10.1109/IROS.2016.7759158}.

\bibitem[Ghalamzan~E. et~al.(2017)Ghalamzan~E., Mavrakis, and
  Stolkin]{ghalamzan2018postgrasp}
A.~M. Ghalamzan~E., N.~Mavrakis, and R.~Stolkin.
\newblock Grasp that optimises objectives along post-grasp trajectories.
\newblock In \emph{2017 5th RSI International Conference on Robotics and
  Mechatronics (ICRoM)}, pages 51--56, 2017.
\newblock \doi{10.1109/ICROM.2017.8466161}.

\bibitem[Han et~al.(2026)Han, Qiu, Bai, Xiao, Zeng, Liu, Yang, Jain, Ma, Fu,
  Zheng, Natarajan, Irshad, Shaw, Gombolay, Kira, and
  Ravichandar]{han2026video2sim2real}
Y.~Han, J.~Qiu, L.~Bai, Z.~Xiao, Z.~Zeng, Y.~Liu, Z.~Yang, S.~Jain, W.~Ma,
  J.~Fu, Y.~Zheng, M.~Natarajan, M.~Z. Irshad, K.~Shaw, M.~Gombolay, Z.~Kira,
  and H.~Ravichandar.
\newblock {Video2Sim2Real}: Full-stack autonomous dexterous skill acquisition
  from a single human video, 2026.
\newblock arXiv:2606.08828.

\bibitem[He et~al.(2024)He, Luo, Xiao, Zhang, Kitani, Liu, and Shi]{he2024h2o}
T.~He, Z.~Luo, W.~Xiao, C.~Zhang, K.~Kitani, C.~Liu, and G.~Shi.
\newblock Learning human-to-humanoid real-time whole-body teleoperation.
\newblock In \emph{IEEE/RSJ International Conference on Intelligent Robots and
  Systems (IROS)}, pages 8944--8951, 2024.
\newblock \doi{10.1109/IROS58592.2024.10801984}.

\bibitem[Heppert et~al.(2024)Heppert, Argus, Welschehold, Brox, and
  Valada]{heppert2024ditto}
N.~Heppert, M.~Argus, T.~Welschehold, T.~Brox, and A.~Valada.
\newblock {DITTO}: Demonstration imitation by trajectory transformation.
\newblock In \emph{IEEE/RSJ International Conference on Intelligent Robots and
  Systems (IROS)}, 2024.
\newblock arXiv:2403.15203.

\bibitem[Hsu et~al.(2025)Hsu, Wen, Xu, Narang, Wang, Zhu, Biswas, and
  Birchfield]{hsu2024spot}
C.-C. Hsu, B.~Wen, J.~Xu, Y.~Narang, X.~Wang, Y.~Zhu, J.~Biswas, and
  S.~Birchfield.
\newblock {SPOT}: {SE(3)} pose trajectory diffusion for object-centric
  manipulation.
\newblock In \emph{IEEE International Conference on Robotics and Automation
  (ICRA)}, 2025.
\newblock arXiv:2411.00965.

\bibitem[Huang et~al.(2026)Huang, Duan, Teng, and Yang]{huang2026physv2a}
H.~Huang, J.~Duan, T.~Teng, and C.~Yang.
\newblock {PhysV2A}: Reachability-gated and semantic-mask-constrained
  feasibility completion for video-to-robot manipulation, 2026.
\newblock arXiv:2607.09365.

\bibitem[Ji et~al.(2026)Ji, Peng, Liu, Li, Yang, Cheng, and
  Wang]{ji2024exbody2}
M.~Ji, X.~Peng, F.~Liu, J.~Li, G.~Yang, X.~Cheng, and X.~Wang.
\newblock {ExBody2}: Advanced expressive humanoid whole-body control.
\newblock In \emph{2026 IEEE International Conference on Robotics and
  Automation (ICRA)}, pages 613--620, 2026.
\newblock \doi{10.1109/ICRA57385.2026.11696204}.

\bibitem[Jiang et~al.(2025)Jiang, Xie, Lin, Xu, Wan, Mandlekar, Fan, and
  Zhu]{jiang2024dexmimicgen}
Z.~Jiang, Y.~Xie, K.~Lin, Z.~Xu, W.~Wan, A.~Mandlekar, L.~Fan, and Y.~Zhu.
\newblock {DexMimicGen}: Automated data generation for bimanual dexterous
  manipulation via imitation learning.
\newblock In \emph{IEEE International Conference on Robotics and Automation
  (ICRA)}, 2025.
\newblock arXiv:2410.24185.

\bibitem[Kareer et~al.(2025)Kareer, Patel, Punamiya, Mathur, Cheng, Wang,
  Hoffman, and Xu]{kareer2024egomimic}
S.~Kareer, D.~Patel, R.~Punamiya, P.~Mathur, S.~Cheng, C.~Wang, J.~Hoffman, and
  D.~Xu.
\newblock {EgoMimic}: Scaling imitation learning via egocentric video.
\newblock In \emph{IEEE International Conference on Robotics and Automation
  (ICRA)}, 2025.
\newblock arXiv:2410.24221.

\bibitem[Kerr et~al.(2024)Kerr, Kim, Wu, Yi, Wang, Goldberg, and
  Kanazawa]{kerr2024rsrd}
J.~Kerr, C.~M. Kim, M.~Wu, B.~Yi, Q.~Wang, K.~Goldberg, and A.~Kanazawa.
\newblock Robot see robot do: Imitating articulated object manipulation with
  monocular {4D} reconstruction.
\newblock In \emph{Conference on Robot Learning (CoRL)}, 2024.
\newblock arXiv:2409.18121.

\bibitem[Khazatsky et~al.(2024)Khazatsky, Pertsch, et~al.]{khazatsky2024droid}
A.~Khazatsky, K.~Pertsch, et~al.
\newblock {DROID}: A large-scale in-the-wild robot manipulation dataset.
\newblock In \emph{Robotics: Science and Systems (RSS)}, 2024.
\newblock arXiv:2403.12945.

\bibitem[Li et~al.(2024)Li, Zhu, Xie, Jiang, Seo, Pavlakos, and
  Zhu]{li2024okami}
J.~Li, Y.~Zhu, Y.~Xie, Z.~Jiang, M.~Seo, G.~Pavlakos, and Y.~Zhu.
\newblock {OKAMI}: Teaching humanoid robots manipulation skills through single
  video imitation.
\newblock In \emph{Conference on Robot Learning (CoRL)}, 2024.
\newblock arXiv:2410.11792.

\bibitem[Liu et~al.(2026)Liu, Patamia, Champion, Dazeley, and
  Cosgun]{liu2026executionaware}
T.~Liu, R.~A. Patamia, B.~Champion, R.~Dazeley, and A.~Cosgun.
\newblock Execution-aware pre-execution ranking for grasp-conditioned robotic
  placement, 2026.
\newblock arXiv:2609.19946.

\bibitem[Lugaresi et~al.(2019)Lugaresi, Tang, Nash, McClanahan, Uboweja, Hays,
  Zhang, Chang, Yong, Lee, Chang, Hua, Georg, and
  Grundmann]{lugaresi2019mediapipe}
C.~Lugaresi, J.~Tang, H.~Nash, C.~McClanahan, E.~Uboweja, M.~Hays, F.~Zhang,
  C.-L. Chang, M.~G. Yong, J.~Lee, W.-T. Chang, W.~Hua, M.~Georg, and
  M.~Grundmann.
\newblock {MediaPipe}: A framework for building perception pipelines, 2019.
\newblock arXiv:1906.08172.

\bibitem[Lum et~al.(2025)Lum, Lee, Liu, and Bohg]{lum2025human2sim2robot}
T.~G.~W. Lum, O.~Y. Lee, C.~K. Liu, and J.~Bohg.
\newblock Crossing the human-robot embodiment gap with sim-to-real {RL} using
  one human demonstration.
\newblock In \emph{Conference on Robot Learning (CoRL)}, volume 305 of
  \emph{Proceedings of Machine Learning Research}, pages 4418--4441, 2025.
\newblock arXiv:2504.12609.

\bibitem[Luo et~al.(2026)Luo, Wang, Zhang, Zheng, Xi, Xu, Xu, Yuan, Zhang,
  Wang, Feng, and Lu]{luo2026beingh05}
H.~Luo, Y.~Wang, W.~Zhang, S.~Zheng, Z.~Xi, C.~Xu, H.~Xu, H.~Yuan, C.~Zhang,
  Y.~Wang, Y.~Feng, and Z.~Lu.
\newblock {Being-H0.5}: Scaling human-centric robot learning for
  cross-embodiment generalization, 2026.
\newblock arXiv:2601.12993.

\bibitem[Mahler et~al.(2017)Mahler, Liang, Niyaz, Laskey, Doan, Liu, Ojea, and
  Goldberg]{mahler2017dexnet}
J.~Mahler, J.~Liang, S.~Niyaz, M.~Laskey, R.~Doan, X.~Liu, J.~A. Ojea, and
  K.~Goldberg.
\newblock {Dex-Net 2.0}: Deep learning to plan robust grasps with synthetic
  point clouds and analytic grasp metrics.
\newblock In \emph{Robotics: Science and Systems (RSS)}, 2017.

\bibitem[Mandlekar et~al.(2023)Mandlekar, Nasiriany, Wen, Akinola, Narang, Fan,
  Zhu, and Fox]{mandlekar2023mimicgen}
A.~Mandlekar, S.~Nasiriany, B.~Wen, I.~Akinola, Y.~Narang, L.~Fan, Y.~Zhu, and
  D.~Fox.
\newblock {MimicGen}: A data generation system for scalable robot learning
  using human demonstrations.
\newblock In \emph{Conference on Robot Learning (CoRL)}, 2023.

\bibitem[Palanisamy et~al.(2026)Palanisamy, Anand, Rathore, Patnaik, Khatana,
  and Janweja]{palanisamy2026mobileego}
S.~Palanisamy, A.~Anand, S.~S. Rathore, P.~Patnaik, S.~Khatana, and E.~Janweja.
\newblock {MobileEgo Anywhere}: Open infrastructure for long horizon egocentric
  data on commodity hardware, 2026.
\newblock arXiv:2605.05945.

\bibitem[Patel et~al.(2026{\natexlab{a}})Patel, Pekarek, Castro~Hernandez, and
  Song]{patel2026bpp}
A.~Patel, B.~Pekarek, J.~E. Castro~Hernandez, and S.~Song.
\newblock What enables in-context behavior prompting for manipulation?,
  2026{\natexlab{a}}.
\newblock arXiv:2606.30457.

\bibitem[Patel et~al.(2026{\natexlab{b}})Patel, Mohan, Mai, Jain, Lazebnik, and
  Li]{patel2026rigvid}
S.~Patel, S.~Mohan, H.~Mai, U.~Jain, S.~Lazebnik, and Y.~Li.
\newblock Robotic manipulation by imitating generated videos without physical
  demonstrations.
\newblock In \emph{International Conference on Learning Representations
  (ICLR)}, 2026{\natexlab{b}}.
\newblock arXiv:2507.00990.

\bibitem[Pavlakos et~al.(2024)Pavlakos, Shan, Radosavovic, Kanazawa, Fouhey,
  and Malik]{pavlakos2024hamer}
G.~Pavlakos, D.~Shan, I.~Radosavovic, A.~Kanazawa, D.~Fouhey, and J.~Malik.
\newblock Reconstructing hands in {3D} with transformers.
\newblock In \emph{IEEE/CVF Conference on Computer Vision and Pattern
  Recognition (CVPR)}, 2024.

\bibitem[Piseno et~al.(2026)Piseno, Tevet, and Liu]{piseno2026cloak}
M.~Piseno, G.~Tevet, and C.~K. Liu.
\newblock Cloak: Zero-shot cross-embodiment manipulation by masking the
  end-effector from the {VLA}, 2026.
\newblock arXiv:2606.22836.

\bibitem[Porges et~al.(2014)Porges, Stouraitis, Borst, and
  Roa]{porges2014reachability}
O.~Porges, T.~Stouraitis, C.~Borst, and M.~A. Roa.
\newblock Reachability and capability analysis for manipulation tasks.
\newblock In \emph{ROBOT2013: First Iberian Robotics Conference}, volume 253 of
  \emph{Advances in Intelligent Systems and Computing}, pages 703--718.
  Springer, 2014.
\newblock \doi{10.1007/978-3-319-03653-3_50}.

\bibitem[Ravi et~al.(2025)Ravi, Gabeur, Hu, Hu, Ryali, Ma, Khedr, R{\"a}dle,
  Rolland, Gustafson, Mintun, Pan, Alwala, Carion, Wu, Girshick, Doll{\'a}r,
  and Feichtenhofer]{ravi2024sam2}
N.~Ravi, V.~Gabeur, Y.-T. Hu, R.~Hu, C.~Ryali, T.~Ma, H.~Khedr, R.~R{\"a}dle,
  C.~Rolland, L.~Gustafson, E.~Mintun, J.~Pan, K.~V. Alwala, N.~Carion, C.-Y.
  Wu, R.~Girshick, P.~Doll{\'a}r, and C.~Feichtenhofer.
\newblock {SAM 2}: Segment anything in images and videos.
\newblock In \emph{International Conference on Learning Representations
  (ICLR)}, 2025.
\newblock arXiv:2408.00714.

\bibitem[Shi et~al.(2026)Shi, Wen, Han, Guo, Hu, Welte, Peng, Stiefelhagen, and
  Rayyes]{shi2026howtransfer}
Y.~Shi, D.~Wen, Z.~Han, Z.~Guo, Y.~Hu, E.~Welte, K.~Peng, R.~Stiefelhagen, and
  R.~Rayyes.
\newblock Hand-centric human-to-robot trajectory transfer from video
  demonstrations via object category-agnostic temporal localization, 2026.
\newblock arXiv:2606.10743.

\bibitem[Sim\'{e}on et~al.(2004)Sim\'{e}on, Laumond, Cort\'{e}s, and
  Sahbani]{simeon2004manipulation}
T.~Sim\'{e}on, J.-P. Laumond, J.~Cort\'{e}s, and A.~Sahbani.
\newblock Manipulation planning with probabilistic roadmaps.
\newblock \emph{The International Journal of Robotics Research}, 23\penalty0
  (7--8):\penalty0 729--746, 2004.
\newblock \doi{10.1177/0278364904045471}.

\bibitem[Trupin et~al.(2025)Trupin, Wang, and Qureshi]{trupin2025itup}
N.~Trupin, Z.~Wang, and A.~H. Qureshi.
\newblock Physics-conditioned grasping for stable tool use, 2025.
\newblock arXiv:2505.01399.

\bibitem[Vahrenkamp et~al.(2013)Vahrenkamp, Asfour, and
  Dillmann]{vahrenkamp2013placement}
N.~Vahrenkamp, T.~Asfour, and R.~Dillmann.
\newblock Robot placement based on reachability inversion.
\newblock In \emph{IEEE International Conference on Robotics and Automation
  (ICRA)}, pages 1970--1975, 2013.
\newblock \doi{10.1109/ICRA.2013.6630839}.

\bibitem[Wang et~al.(2024)Wang, Shi, Wang, Zhang, Fei-Fei, and
  Liu]{wang2024dexcap}
C.~Wang, H.~Shi, W.~Wang, R.~Zhang, L.~Fei-Fei, and C.~K. Liu.
\newblock {DexCap}: Scalable and portable mocap data collection system for
  dexterous manipulation.
\newblock In \emph{Robotics: Science and Systems (RSS)}, 2024.

\bibitem[Wang et~al.(2026{\natexlab{a}})Wang, Ren, Morgan, Chen, Qian,
  Chanrungmaneekul, and Hang]{wang2026egoinfinity}
G.~Wang, K.~Ren, A.~Morgan, Y.~Chen, H.~H. Qian, P.~Chanrungmaneekul, and
  K.~Hang.
\newblock {EgoInfinity}: A web-scale {4D} hand-object interaction data engine
  for any-view robot retargeting and video-to-action robot learning,
  2026{\natexlab{a}}.
\newblock arXiv:2606.17385.

\bibitem[Wang et~al.(2020)Wang, Xiang, and Fox]{wang2020manipulationtrajopt}
L.~Wang, Y.~Xiang, and D.~Fox.
\newblock Manipulation trajectory optimization with online grasp synthesis and
  selection.
\newblock In \emph{Robotics: Science and Systems (RSS)}, 2020.

\bibitem[Wang et~al.(2026{\natexlab{b}})Wang, Lin, Chen, Yuan, Liang, Huang,
  Chen, Lei, Zhang, Zhang, Li, Zhang, Xiao, Jiao, and Jin]{wang2026ego2robot}
Y.~Wang, P.~Lin, X.-H. Chen, H.~Yuan, Z.~Liang, Y.~Huang, A.~Chen, Z.~Lei,
  J.~Zhang, T.~Zhang, H.~Li, T.~Zhang, C.~Xiao, Z.~Jiao, and Q.~Jin.
\newblock {Ego2Robot}: Scalable robot data synthesis from egocentric human
  data, 2026{\natexlab{b}}.
\newblock arXiv:2608.02580.

\bibitem[Wen et~al.(2024)Wen, Lin, So, Chen, Dou, Gao, and Abbeel]{wen2024atm}
C.~Wen, X.~Lin, J.~So, K.~Chen, Q.~Dou, Y.~Gao, and P.~Abbeel.
\newblock Any-point trajectory modeling for policy learning.
\newblock In \emph{Robotics: Science and Systems (RSS)}, 2024.
\newblock arXiv:2401.00025.

\bibitem[Xiang et~al.(2025)Xiang, Lv, Xu, Deng, Wang, Zhang, Chen, Tong, and
  Yang]{xiang2025trellis}
J.~Xiang, Z.~Lv, S.~Xu, Y.~Deng, R.~Wang, B.~Zhang, D.~Chen, X.~Tong, and
  J.~Yang.
\newblock Structured {3D} latents for scalable and versatile {3D} generation.
\newblock In \emph{IEEE/CVF Conference on Computer Vision and Pattern
  Recognition (CVPR)}, 2025.

\bibitem[Xu et~al.(2024)Xu, Xu, Xu, Chi, Wetzstein, Veloso, and
  Song]{xu2024im2flow2act}
M.~Xu, Z.~Xu, Y.~Xu, C.~Chi, G.~Wetzstein, M.~Veloso, and S.~Song.
\newblock Flow as the cross-domain manipulation interface.
\newblock In \emph{Conference on Robot Learning (CoRL)}, 2024.
\newblock arXiv:2407.15208.

\bibitem[Xu et~al.(2026)Xu, Park, Zhang, Cousineau, Bhat, Barreiros, Wang,
  Bohg, and Song]{xu2026hommi}
X.~Xu, J.~Park, H.~Zhang, E.~Cousineau, A.~Bhat, J.~Barreiros, D.~Wang,
  J.~Bohg, and S.~Song.
\newblock {HoMMI}: Learning whole-body mobile manipulation from human
  demonstrations.
\newblock In \emph{Robotics: Science and Systems (RSS)}, Sydney, Australia,
  2026.
\newblock \doi{10.15607/RSS.2026.XXII.205}.

\bibitem[Xue and Dillmann(2011)]{xue2011reachability}
Z.~Xue and R.~Dillmann.
\newblock Efficient grasp planning with reachability analysis.
\newblock \emph{International Journal of Humanoid Robotics}, 8\penalty0
  (4):\penalty0 761--775, 2011.
\newblock \doi{10.1142/S0219843611002654}.

\bibitem[Ye et~al.(2025)Ye, Liu, Ding, Gao, Rybkin, and
  Abbeel]{ye2025video2policy}
W.~Ye, F.~Liu, Z.~Ding, Y.~Gao, O.~Rybkin, and P.~Abbeel.
\newblock {Video2Policy}: Scaling up manipulation tasks in simulation through
  internet videos, 2025.
\newblock arXiv:2502.09886.

\bibitem[Yin et~al.(2026)Yin, Li, Gui, and Liu]{yin2026feasiblecap}
Z.~Yin, F.~Li, Y.~Gui, and J.~Liu.
\newblock {FeasibleCap}: Real-time embodiment constraint guidance for
  in-the-wild robot demonstration collection, 2026.
\newblock arXiv:2603.07580.

\bibitem[Yu et~al.(2025)Yu, Fu, Huang, El-Refai, Ambrus, Cheng, Irshad, and
  Goldberg]{yu2025real2render2real}
J.~Yu, L.~Fu, H.~Huang, K.~El-Refai, R.~A. Ambrus, R.~Cheng, M.~Z. Irshad, and
  K.~Goldberg.
\newblock {Real2Render2Real}: Scaling robot data without dynamics simulation or
  robot hardware.
\newblock In \emph{Conference on Robot Learning (CoRL)}, volume 305 of
  \emph{Proceedings of Machine Learning Research}, pages 547--577, 2025.
\newblock arXiv:2505.09601.

\bibitem[Yu and Qiu(2026)]{yu2026so101bench}
Y.~Yu and X.~Qiu.
\newblock Benchmarking vision-language-action models on {SO-101}: Failure and
  recovery analysis, 2026.
\newblock arXiv:2606.08881.

\bibitem[Yuan et~al.(2024)Yuan, Wen, Zhang, and Gao]{yuan2024generalflow}
C.~Yuan, C.~Wen, T.~Zhang, and Y.~Gao.
\newblock General flow as foundation affordance for scalable robot learning.
\newblock In \emph{Conference on Robot Learning (CoRL)}, 2024.
\newblock arXiv:2401.11439.

\bibitem[Zacharias et~al.(2007)Zacharias, Borst, and
  Hirzinger]{zacharias2007capability}
F.~Zacharias, C.~Borst, and G.~Hirzinger.
\newblock Capturing robot workspace structure: representing robot capabilities.
\newblock In \emph{IEEE/RSJ International Conference on Intelligent Robots and
  Systems (IROS)}, pages 3229--3236, 2007.
\newblock \doi{10.1109/IROS.2007.4399105}.

\bibitem[Zhai et~al.(2026)Zhai, Zeng, Lu, Farhadi, Wang, and Ma]{zhai2026psi}
A.~J. Zhai, K.-H. Zeng, J.~Lu, A.~Farhadi, S.~Wang, and W.-C. Ma.
\newblock Imitating what works: Simulation-filtered modular policy learning
  from human videos.
\newblock \emph{Transactions on Machine Learning Research (TMLR)}, 2026.
\newblock arXiv:2602.13197.

\bibitem[Zhu et~al.(2026)Zhu, Lim, Stone, and Zhu]{zhu2024orion}
Y.~Zhu, A.~Lim, P.~Stone, and Y.~Zhu.
\newblock Vision-based manipulation from single human video with open-world
  object graphs.
\newblock \emph{Autonomous Robots}, 50\penalty0 (2):\penalty0 27, 2026.
\newblock \doi{10.1007/s10514-026-10253-8}.

\end{thebibliography}

\end{document}